\documentclass[11pt]{article}

\PassOptionsToPackage{numbers, compress}{natbib}
\usepackage{natbib}

\usepackage[final]{neurips_2020}

\usepackage[utf8]{inputenc} 
\usepackage[T1]{fontenc}    
\usepackage{hyperref}       
\usepackage{url}            
\usepackage{booktabs}       
\usepackage{amsfonts}       
\usepackage{nicefrac}       
\usepackage{microtype}      

\usepackage{graphicx}

\usepackage{amsmath,amsfonts,bm}

\def\Figref#1{Figure~\ref{#1}}

\def\Secref#1{Section~\ref{#1}}

\def\eqref#1{equation~\ref{#1}}
\def\Eqref#1{Equation~\ref{#1}}

\def\1{\bm{1}}

\def\eps{{\epsilon}}

\def\ve{{\bm{e}}}

\def\vo{{\bm{o}}}
\def\vp{{\bm{p}}}
\def\vq{{\bm{q}}}

\def\vw{{\bm{w}}}
\def\vx{{\bm{x}}}
\def\vy{{\bm{y}}}
\def\vz{{\bm{z}}}
\def\vpi{{\bm{\pi}}}
\def\vphi{{\bm{\phi}}}
\def\vsigma{{\bm{\sigma}}}
\def\vrho{{\bm{\rho}}}
\def\iu{{\mathrm{i}}}

\def\mA{{\bm{A}}}
\def\mB{{\bm{B}}}
\def\mC{{\bm{C}}}
\def\mD{{\bm{D}}}
\def\mE{{\bm{E}}}

\def\mH{{\bm{H}}}
\def\mI{{\bm{I}}}

\def\mM{{\bm{M}}}

\def\mP{{\bm{P}}}
\def\mQ{{\bm{Q}}}

\def\mX{{\bm{X}}}
\def\mY{{\bm{Y}}}
\def\mZ{{\bm{Z}}}

\def\mPi{{\bm{\Pi}}}
\DeclareMathAlphabet{\mathsfit}{\encodingdefault}{\sfdefault}{m}{sl}
\SetMathAlphabet{\mathsfit}{bold}{\encodingdefault}{\sfdefault}{bx}{n}

\def\sC{{\mathbb{C}}}

\newcommand{\E}{\mathbb{E}}

\newcommand{\R}{\mathbb{R}}

\DeclareMathOperator{\Tr}{Tr}

\newcommand{\fourEEzz}{\,\vcenter{\hbox{\ensuremath{%
  \mathchoice{\includegraphics[height=2.5ex]{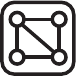}}
    {\includegraphics[height=2.5ex]{figs/MiniFig_4_3322}}
    {\includegraphics[height=1.5ex]{figs/MiniFig_4_3322}}
    {\includegraphics[height=1ex]{figs/MiniFig_4_3322}}
}}}\,}
\usepackage{amssymb}
\usepackage{amsfonts}
\usepackage{amsthm}
\usepackage{hyperref}
\usepackage{url}
\usepackage{booktabs}
\usepackage[linesnumbered,boxed,ruled,vlined]{algorithm2e}

\usepackage{xcolor}
\usepackage{subfig}
\usepackage{thmtools,thm-restate}
\usepackage{amsmath}
\usepackage{mathtools}
\usepackage{enumitem}
\usepackage{wrapfig}
\newcommand{\norm}[2]{\left\lVert#1\right\rVert_{#2}}

\DeclareMathOperator{\vect}{vec}
\newtheorem{definition}{Definition}
\newtheorem{lemma}{Lemma}

\newtheorem{remark}{Remark}
\newtheorem{claim}{Claim}

\newcommand{\comment}[1]{} 
\newcommand{\vpara}[1]{\vspace{0.2cm}\noindent\textbf{#1 }}

\newcommand{\beq}[1]{{\small \begin{equation}#1\end{equation}
}}

\newcommand{\besp}[1]{\begin{split}#1\end{split}}

\DeclarePairedDelimiter{\diagfences}{(}{)}
\newcommand{\diag}{\operatorname{diag}\diagfences}

\newcommand{\abs}[1]{\left\lvert#1\right\rvert}
\DeclareMathOperator{\vol}{vol}

\newcommand{\jiezhong}[1]{\textcolor{green}{(Jiezhong: {#1})}} 

\title{
  A Matrix Chernoff Bound for Markov Chains and Its Application to Co-occurrence Matrices
}

\author{%
Jiezhong Qiu\\
Tsinghua University\\
\texttt{qiujz16@mails.tsinghua.edu.cn}\\
\And
Chi Wang\\
Microsoft Research, Redmond\\
\texttt{wang.chi@microsoft.com}\\
\And
Ben Liao\\
Tencent Quantum Lab\\
\texttt{bliao@tencent.com}\\
\And
Richard Peng\\
Georgia Tech\\
\texttt{rpeng@cc.gatech.edu}\\
\And
Jie Tang\\
Tsinghua University\\
\texttt{jietang@tsinghua.edu.cn}
}

\comment{
\author{%
Jiezhong Qiu\\
Tsinghua University\\
\texttt{qiujz16@mails.tsinghua.edu.cn}\\
\and
Chi Wang\\
Microsoft Research, Redmond\\
\texttt{wang.chi@microsoft.com}\\
\and
Ben Liao\\
Tencent\\
\texttt{bliao@tencent.com}\\
\and
Richard Peng\\
Georgia Tech\\
\texttt{rpeng@cc.gatech.edu}\\
\and
Jie Tang\\
Tsinghua University\\
\texttt{jietang@tsinghua.edu.cn}
}
}

\begin{document}

\maketitle

\begin{abstract}

\comment{
Co-occurrence statistics for sequential data are common and important
data signals in machine learning, which provide rich correlation and clustering information about the underlying object space.
We give the first bound on the convergence rate of 
estimating the co-occurrence matrix of a regular~(aperiodic and irreducible) finite Markov
chain from a single random walk trajectory. 
Our work is motivated by the analysis of a well-known graph learning algorithm DeepWalk by
[Qiu et al. WSDM '18], 
who study the convergence~(in probability) of
co-occurrence matrix from random walk on undirected  graphs in the limit,
but left the convergence rate an open problem. 

We prove a Chernoff-type bound for sums of matrix-valued random variables sampled via a regular Markov chain, generalizing
the regular undirected graph case studied by [Garg et al. STOC '18].
Using the Chernoff-type bound, we show that given a regular Markov chain with $n$ states and mixing time $\tau$, 
we need a trajectory of length $O(\tau (\log{n} + \log{\tau})/\eps^2)$ to achieve an estimator of the co-occurrence matrix with error bound $\epsilon$.
We conduct several experiments and the experimental results are consistent with the exponentially fast convergence rate from 
theoretical analysis. 
Our result gives the first sample complexity analysis in graph representation learning. 
}

We prove a Chernoff-type bound for sums of matrix-valued random variables 
sampled via a regular (aperiodic and irreducible) finite Markov chain.
Specially, consider a random walk on a regular Markov chain and a Hermitian matrix-valued function on its state space. Our result gives exponentially decreasing bounds on the tail distributions of the  extreme eigenvalues of the sample mean matrix.
Our proof is based on  the 
matrix expander (regular undirected graph) Chernoff bound [Garg et al. STOC '18] and scalar
Chernoff-Hoeffding bounds for  Markov
chains [Chung et al. STACS '12].

Our matrix Chernoff bound for Markov chains can be applied 
to analyze the behavior of co-occurrence statistics for sequential data,
which have been  common and important
data signals in machine learning. 
We show that given a regular Markov chain with $n$ states and mixing time $\tau$, 
we need a trajectory of length $O(\tau (\log{n} + \log{\tau})/\eps^2)$ to achieve an estimator of the co-occurrence matrix with error bound $\epsilon$.
We conduct several experiments and the experimental results are consistent with the exponentially fast convergence rate from 
theoretical analysis. 
Our result gives the
first bound on the convergence rate of  the co-occurrence matrix and
the first sample complexity analysis in graph representation learning. 
\end{abstract}
\section{Introduction}



Chernoff bound~\cite{chernoff1952measure},
which gives exponentially decreasing bounds on tail distributions of sums of independent scalar-valued random variables,
is one of the most basic and versatile tools in theoretical computer science, with countless applications to practical problems~\cite{kearns1994introduction, motwani1995randomized}.
There are two notable limitations when applying Chernoff bound in analyzing sample complexity in real-world machine learning problems. First, in many cases the random variables have dependence, e.g., Markov dependence~\cite{karlin2014first} in MCMC~\cite{jerrum1996markov} and online learning~\cite{tekin2010online}. Second, applications are often concerned with the concentration behavior of quantities beyond scalar-valued random variables, e.g.,
random features in kernel machines~\cite{rahimi2008random} and co-occurrence statistics which are random matrices~\cite{perozzi2014deepwalk,qiu2018network}.

Existing research has attempted to extend the original Chernoff bound in one of these two limitations~\cite{kahale1997large,gillman1998chernoff,lezaud1998chernoff,leon2004optimal,wigderson2005randomness,healy2008randomness,chung2012chernoff,rao2017sharp,wagner2008tail,rudelson1999random,ahlswede2002strong,tropp2015introduction}. \citet{wigderson2005randomness} conjectured that 
Chernoff bounds can be generalized to both matrix-valued random variables 
and Markov dependence, while restricting the Markov dependence to be \emph{a regular undirected graph}.
It was recently proved by \citet{garg2018matrix},
based on  a new multi-matrix extension of the Golden-Thompson inequality~\cite{sutter2017multivariate}.
However, the  regular undirected graph is a special case of  Markov chains which are reversible and have a uniform stationary distribution, and does not apply to practical problems such as random walk on generic graphs. It is an open question for the Chernoff bound of matrix-valued random matrices with more generic Markov dependence.

In this work, we establish large deviation bounds
for the tail probabilities of the extreme eigenvalues of sums of random matrices
sampled via a regular Markov chain\footnote{Please note that 
regular Markov chains are Markov chains which are aperiodic and irreducible, 
while an undirected regular graph is an undirected graph where each vertex has the same number of neighbors. 
In this work, the term ``regular'' may  have different meanings depending on the context.} starting from an arbitrary distribution~(not necessarily the stationary distribution), which significantly improves the result of \citet{garg2018matrix}. 
\comment{
We consider a regular Markov chain $\mP$ with stationary distribution $\vpi$ and a bounded Hermitian matrix-valued function 
$f: [N]\rightarrow \mathbb{C}^{n\times n}$ on the state space $[N]$ 
with  $\sum \pi_v f(v) = 0$.
We show that, for a $k$-step random walk $(v_1,\cdots, v_k)$ on $\mP$, 
the probability that the extreme eigenvalues of its sample mean 
deviating from zero by $\epsilon$ is at most $d \exp{\left(-\Omega\left(\eps^2(1-\lambda)k\right)\right)}$, where $\lambda$ is
the spectral expansion of $\mP$.
}
More formally, we prove the following theorem:
\begin{restatable}[Markov Chain Matrix Chernoff Bound]{theorem}{complexchernoff}
    \label{thm:complexchernoff}
    Let $\mP$ be a regular Markov chain with state space  $[N]$,
    stationary distribution $\vpi$ and spectral expansion $\lambda$.
    Let $f: [N] \rightarrow \mathbb{C}^{d\times d}$
    be a function such that (1) $\forall v \in [N]$, $f(v)$ is Hermitian and $\norm{f(v)}{2}\leq 1$; (2) $\sum_{v\in [N]} \pi_v f(v) = 0$.
    Let $(v_1,\cdots, v_k)$ denote a $k$-step random walk on $\mP$ starting from a distribution $\vphi$. Given $\eps\in (0, 1)$,
    \beq{
        \nonumber
        \besp{
            \mathbb{P}\left[\lambda_{\max}\left( \frac{1}{k}\sum_{j=1}^k f(v_j)\right)\geq \epsilon\right] &\leq 4\norm{\vphi}{\vpi}d^{2}\exp{\left( -(\eps^2 (1-\lambda)k / 72) \right)} \\
            \mathbb{P}\left[\lambda_{\min}\left( \frac{1}{k}\sum_{j=1}^k f(v_j)\right)\leq -\epsilon\right] &\leq 4\norm{\vphi}{\vpi}d^{2} \exp{\left( - (\eps^2 (1-\lambda)k / 72) \right)}.
        }}
\end{restatable}
In the above theorem, $\norm{\cdot}{\vpi}$ is the $\vpi$-norm~(which we define formally later in \Secref{sec:preliminaries}) measuring
the distance between the initial distribution $\vphi$ and the stationary distribution $\vpi$.
Our strategy is to incorporate the
concentration of matrix-valued functions from~\cite{garg2018matrix} into the study
of general Markov chains from~\cite{chung2012chernoff},
which was originally for scalars.
\comment{
Key to this extension is the definition of an inner product related to the stationary distribution $\vpi$ of Markov chain $\mP$, 
and a spectral expansion from such inner products.
In contrast, the regular undirected  graph case studied in \cite{garg2018matrix} can be handled using the standard
inner products, as well as the second largest absolute eigenvalues
of $\mP$ instead of the spectral expansion.
}


\subsection{Applications to Co-occurrence Matrices of Markov Chains}

\begin{wrapfigure}{l}{0.455\textwidth}
\begin{minipage}{0.455\textwidth}
\begin{algorithm}[H]
    \small
    \SetCustomAlgoRuledWidth{0.45\textwidth}
    \caption{The Co-occurrence Matrix.}
    \label{alg:dw}
    \textbf{Input}  sequence $(v_1, \cdots, v_L)$; window size $T$\; 
    \textbf{Output} co-occurrence matrix $\mC$\;
    $\mC \gets \bm{0}_{n\times n}$; \tcc*{$v_i \in [n], i\in [L]$}
    \For{$i = 1,2,\ldots, L-T$}{
    \For{$r = 1, \ldots, T$} {
        $\mC_{v_i, v_{i+r}} \gets \mC_{v_i, v_{i+r}} + 1/T$\; 
        $\mC_{v_{i+r}, v_i} \gets \mC_{v_{i+r}, v_i} + 1/T$\; 
    }}
    $\mC \gets \frac{1}{2(L-T)} \mC $\;
    \textbf{Return}  $\mC$\;  
    \normalsize
\end{algorithm}
\end{minipage}
\end{wrapfigure}

%
The co-occurrence statistics have recently emerged as common 
and important data signals in
machine learning, providing rich correlation and clustering information about the underlying object space,
such as the word co-occurrence in natural language processing~\cite{mikolov2013efficient, NIPS2013_5021, mikolov2013linguistic, NIPS2014_5477,pennington2014glove}, 
vertex co-occurrence in graph learning~\cite{perozzi2014deepwalk,tang2015line,grover2016node2vec,hamilton2017inductive,dong2017metapath2vec,qiu2018network}, 
item co-occurrence in recommendation system~\cite{shani2005mdp,liang2016factorization,barkan2016item2vec,vasile2016meta,liu2017related},
action co-occurrence in reinforcement learning~\cite{tennenholtz2019natural}, and emission co-occurrence of hidden Markov models~\cite{kontorovich2013learning,huang2018learning,mattila2020fast}.
Given a sequence of objects $(v_1, \cdots, v_L)$, the co-occurrence statistics are computed by  moving a sliding window of fixed size $T$  over 
the sequence and recording the frequency of objects' co-occurrence within the sliding window. 
A pseudocode of the above procedure is listed in Algorithm~\ref{alg:dw},
which produces an $n$ by $n$ co-occurrence matrix where $n$ is the size of the object space.

A common assumption when building such co-occurrence matrices is that the sequential data 
are long enough to provide an accurate estimation. 
For instance, \citet{NIPS2013_5021} use a news article dataset 
with one billion words in their Skip-gram model; 
\citet{tennenholtz2019natural} train their Act2vec model with action sequences from over a million StarCraft II game replays, 
which are equivalent to 100 years of consecutive gameplay;
\citet{perozzi2014deepwalk} samples large amounts of random walk sequences from graphs to capture the vertex co-occurrence.
A recent work by \citet{qiu2018network} studies the convergence of co-occurrence matrices of random walk on undirected graphs in the limit~(i.e., when the length of random walk goes to infinity),
but left the convergence rate an open problem.
It remains unknown whether the co-occurrence statistics are sample efficient and how efficient they are.

In this paper, we study the situation where the sequential
data are sampled from
a regular finite Markov chain~(i.e., an aperiodic and irreducible finite Markov chain), 
and derive bounds on the sample efficiency of  co-occurrence matrix estimation, specifically on the
length of the trajectory needed in the sampling algorithm shown in Algorithm~\ref{alg:dw}.
To give a formal statement,
we first translate Algorithm~\ref{alg:dw} to linear algebra language. 
Given a trajectory $(v_1, \cdots, v_L)$ from state space $[n]$ and step weight coefficients $(\alpha_1, \cdots, \alpha_T)$,
the co-occurrence matrix is defined to be
\comment{
\beq{\nonumber
    \mC \triangleq \frac{1}{L-T}\sum_{i=1}^{L-T} \sum_{r=1}^T\frac{\alpha_r}{2}\left(\ve_{v_i}  \ve^\top_{v_{i+r}} + \ve_{v_{i+r}} \ve^\top_{v_i}\right).
}where $\ve_{v_i}$ is a length-$n$ vector with a one in its $v_i$-th entry and zeros elsewhere.
\jiezhong{If $(v_1,\cdots, v_L)$ is not a stationary walk, we gonna use the following formulation, by introducing the term ``Asymptotic Expectation''}
}
\beq{\nonumber
    \mC \triangleq \frac{1}{L-T}\sum_{i=1}^{L-T} \mC_i, \text{where } \mC_i \triangleq\sum_{r=1}^T\frac{\alpha_r}{2}\left(\ve_{v_i}  \ve^\top_{v_{i+r}} + \ve_{v_{i+r}} \ve^\top_{v_i}\right).
}Here $\mC_i$ accounts for the co-occurrence within sliding window $(v_i, \cdots, v_{i+T})$, and $\ve_{v_i}$ is a length-$n$ vector with a one in its $v_i$-th entry and zeros elsewhere.
Thus $\small \ve_{v_i}  \ve^\top_{v_{i+r}}\normalsize$ is a $n$ by $n$ matrix with its $(v_i, v_{i+r})$-th entry to be one and other entries to be zero,
which records the co-occurrence of $v_i$ and $v_{i+r}$. 
Note that Algorithm~\ref{alg:dw} is a special case when step weight coefficients are uniform, i.e., $\alpha_r=1/T, r\in[T]$,
and the co-occurrence statistics in all the applications mentioned above can be formalized in this way.
When trajectory $(v_1, \cdots, v_L)$ is a random walk
from a regular Markov chain $\mP$ with stationary distribution $\vpi$,
 the asymptotic expectation of the co-occurrence matrix within sliding window $(v_i, \cdots, v_{i+L})$ is
\beq{\nonumber
\mathbb{AE}[\mC_i]\triangleq \lim_{i\to\infty}\mathbb{E}(\mC_i) = \sum_{r=1}^T\frac{\alpha_r}{2} \left(\mPi \mP^r + \left(\mPi\mP^r\right)^\top \right),
}where $\mPi\triangleq\diag{\vpi}$. Thus the asymptotic expectation of the co-occurrence matrix is
\comment{
\beq{\nonumber
\besp{
\E\left[\ve_{v_i}  \ve^\top_{v_{i+r}}(w, c)\right] &=\text{Prob}[v_i = w, v_{i+r}=c]= \underbrace{\text{Prob}[v_i = w]}_{\text{converges to }  \pi(w)}\underbrace{\text{Prob}[v_{i+r}=c|v_i=w]}_{=\mP^{r}(w, c) } \longrightarrow \pi(w) \mP^{r}(w, c) \\
\lim_{i\to\infty}\E\left[\ve_{v_i}  \ve^\top_{v_{i+r}}\right] &= \mPi \mP^r \text{ where }  \mPi\triangleq\diag{\vpi}
}}
}
\comment{
\beq{\label{eq:expected}
    \mathbb{E}\left[\mC\right] 
    = \frac{1}{L-T}\sum_{i=1}^{L-T} \sum_{r=1}^T\frac{\alpha_r}{2} \left(\mathbb{E}\left[\ve_{v_i}  \ve^\top_{v_{i+r}}\right] + \mathbb{E}\left[\ve_{v_{i+r}} \ve^\top_{v_i}\right] \right) = \sum_{r=1}^T\frac{\alpha_r}{2} \left(\mPi \mP^r + \left(\mPi\mP^r\right)^\top \right),
}
}
\beq{\label{eq:expected}
\mathbb{AE}[\mC]\triangleq \lim_{L\to\infty}\mathbb{E}\left[\mC\right] 
    = \lim_{L\to\infty}\frac{1}{L-T}\sum_{i=1}^{L-T}\mathbb{E}(\mC_i)  = \sum_{r=1}^T\frac{\alpha_r}{2} \left(\mPi \mP^r + \left(\mPi\mP^r\right)^\top \right).
}Our main result regarding the estimation of the co-occurrence
matrix is the following convergence bound related to the length of the walk sampled.
\begin{restatable}[Convergence Rate of Co-occurrence Matrices]{theorem}{rate}
    \label{thm:rate}
    Let $\mP$ be a regular Markov chain with state space  $[n]$,
    stationary distribution $\vpi$ and mixing time $\tau$.
    Let $(v_1,\cdots, v_L)$ denote a $L$-step random walk on $\mP$ starting from a distribution $\vphi$ on $[n]$.
    Given 
    step weight coefficients $(\alpha_1, \cdots, \alpha_T)$ s.t. $\sum_{r=1}^T \abs{\alpha_r}=1$, 
    and $\eps\in (0, 1)$, the probability that the co-occurrence matrix $\mC$ deviates from its asymptotic expectation $\mathbb{AE}[\mC]$~(in 2-norm) is bounded by:
    \beq{\nonumber
    \mathbb{P}\left[\norm{\mC-\mathbb{AE}[\mC]}{2} \geq \eps\right] \leq 2\left(\tau + T\right)\norm{\vphi}{\vpi}  n^{2} \exp{\left( -\frac{\eps^2(L-T)}{576\left(\tau + T\right)} \right)}.
    }Specially, there exists a trajectory length  $L=O\left((\tau+T)(\log{n} + \log{(\tau+T)})/\eps^2 + T\right)$ such that
    $
    \mathbb{P}\left[ \norm{ \mC - \mathbb{AE}[\mC] }{2} \geq \epsilon\right] \leq \frac{1}{n^{O(1)}}
    $. Assuming $T=O(1)$ gives $L=O\left(\tau(\log{n} + \log{\tau})/\eps^2\right)$.
\end{restatable}

Our result in Theorem~\ref{thm:rate} gives the first sample complexity analysis for many graph representation learning algorithms.
Given a graph, these algorithms aim to
learn a function from the vertices to a low dimensional vector space.
Most of them~(e.g., DeepWalk~\cite{perozzi2014deepwalk}, node2vec~\cite{grover2016node2vec},
metapath2vec~\cite{dong2017metapath2vec}, GraphSAGE~\cite{hamilton2017inductive}) consist of two steps.
The first step is to draw random sequences from a stochastic
process defined on the graph and then count co-occurrence statistics from the sampled sequences,
where the stochastic process is usually defined to be first-order
or higher-order random walk on the graph.
The second step is to train a model to fit the co-occurrence statistics.
For example, DeepWalk can be viewed as factorizing a point-wise mutual information matrix~\cite{NIPS2014_5477,qiu2018network}
which is a transformation of the co-occurrence matrix; GraphSAGE fits the co-occurrence statistics
with a graph neural network~\cite{kipf2017semi}.
The common assumption is that there are enough  samples so that the co-occurrence statistics are accurately estimated.
We are the first work to study the sample complexity of the aforementioned algorithms.
Theorem~\ref{thm:rate} implies that these algorithms need $O(\tau (\log{n} + \log{\tau})/\eps^2)$ samples to achieve
a good estimator of the co-occurrence matrix.

\vpara{Previous work}
\citet{hsu2015mixing,hsu2019mixing} study a similar problem. They leverage the  co-occurrence matrix with $T=1$
to estimate the mixing time 
in reversible Markov chains from a single trajectory.
Their main technique is a blocking technique~\cite{yu1994rates} which is in parallel 
with the Markov chain matrix Chernoff-bound used in this work.
Our work is also related to the research  about random-walk matrix polynomial sparsification
when the Markov chain $\mP$ is a random walk on an undirected graph.
In this case, we can rewrite $\mP=\mD^{-1}\mA$ where $\mD$ and $\mA$ 
is the degree matrix and adjacency
matrix of an undirected graph with $n$ vertices and $m$ edges, 
and the expected co-occurrence matrix in \Eqref{eq:expected} can be simplified as
$\mathbb{AE}\left[\mC\right]
    = \frac{1}{\vol{(G)}}\sum_{r=1}^T\alpha_r \mD (\mD^{-1}\mA)^r$,\footnote{The volume of a graph $G$ is defined to be $\vol{(G)} \triangleq \sum_i\sum_j \mA_{ij}$.}
which is known as \emph{random-walk matrix polynomials}~\cite{cheng2015efficient,cheng2015spectral}.
\citet{cheng2015spectral} propose an algorithm which needs $O(T^2m\log{n}/\eps^2)$ steps of random walk to construct
an $\eps$-approximator for the random-walk matrix polynomials. Our bound in
Theorem~\ref{thm:rate} is stronger than the bound proposed by \citet{cheng2015spectral}
when the Markov chain $\mP$ mixes fast. Moreover, \citet{cheng2015spectral} require $\alpha_r$ to be non-negative, while our bound can handle
negative step weight coefficients.

\vpara{Organization} The rest of the paper is organized as follows. 
In \Secref{sec:preliminaries} we provide preliminaries, 
followed by the proof of matrix Chernoff bound in \Secref{sec:chernoff} and the proof of convergence rate of co-occurrence matrices
in \Secref{sec:rate}.
In \Secref{sec:exp}, we conduct experiments on both synthetic and real-world datasets.
Finally, we conclude this work in \Secref{sec:conclusion}.

\section{Preliminaries}
\label{sec:preliminaries}

In this paper, we denote $\mP$ to be a finite Markov chain on $n$ states.
$\mP$ could refer to either the chain itself
or the corresponding transition probability matrix --- an $n$ by $n$ matrix such that its entry
$\mP_{ij}$ indicates the probability that state $i$ moves to state $j$. A Markov chain is 
called an ergodic Markov chain if it is possible to
eventually get from every state to every other state with positive probability.
A Markov chain is regular if some power of its transition matrix has all strictly positive entries.
A regular Markov chain must
    be an ergodic Markov chain, but not vice versa.
    An ergodic Markov chain has unique stationary distribution, i,e., there exists a unique probability vector 
    $\vpi$ such that $\vpi^\top = \vpi^\top \mP$. For convenience, we denote $\mPi \triangleq \diag{\vpi}$. 

    The time that a regular Markov chain\footnote{Please note that we need the Markov chain to be regular to make the mixing-time 
    well-defined. For an ergodic Markov chain which could be periodic, the mixing time may be ill-defined.} needs to be ``close'' to its stationary distribution is called \emph{mixing time}. 
    Let $\vx$ and $\vy$ be two probability vectors.
    The \emph{total variation distance} between them is
    $\norm{\vx - \vy}{TV} 
    \triangleq \frac{1}{2} \norm{\vx-\vy}{1}$.
    For $\delta>0$, the $\delta$-mixing time of regular Markov chain $\mP$ is 
    $\small
    \tau(\mP) \triangleq \min\left\{ t: \max_{\vx} \norm{(\vx^\top\mP^t)^\top - \vpi}{TV} \leq \delta\right\}
    \normalsize
    $, where
    $\vx$ is an arbitrary probability vector. 
    
The stationary distribution $\vpi$  also defines 
a inner product space where the inner product~(under $\vpi$-kernel)
is defined as
$\langle \vx, \vy \rangle_{\vpi} \triangleq \vy^\ast \mPi^{-1} \vx$ for $\forall \vx, \vy \in \mathbb{C}^N$, 
where $\vy^\ast$ is the conjugate transpose of $\vy$.
A naturally defined norm based on the above inner product
is $\norm{\vx}{\vpi} \triangleq \sqrt{\langle \vx, \vx \rangle_{\vpi}}$. 
Then we can define the \emph{spectral expansion} $\lambda(\mP)$ of a Markov chain $\mP$~\cite{mihail1989conductance,fill1991eigenvalue, chung2012chernoff} as 
$\small
\lambda(\mP) \triangleq \max_{\langle\vx, \vpi\rangle_{\vpi}=0, \vx\neq 0} \frac{\norm{\left(\vx^\ast \mP\right)^\ast}{\vpi}}{\norm{\vx}{\vpi}}
\normalsize
$.
The spectral expansion $\lambda(\mP)$ is known to be a measure of mixing time of a Markov chain. The smaller $\lambda(\mP)$ is, the faster 
a Markov chain converges to its stationary distribution~\cite{wolfer2019estimating}.
If $\mP$ is reversible, $\lambda(\mP)$ is simply the second largest absolute eigenvalue of $\mP$~(the largest is always $1$).
The irreversible case is more complicated, since $\mP$ may have complex eigenvalues. 
In this case, $\lambda(\mP)$ 
is actually the square root of the second largest absolute eigenvalue of the  
\emph{multiplicative reversiblization} of $\mP$~\cite{fill1991eigenvalue}.
When $\mP$ is clear from the context, we will simply write $\tau$ and $\lambda$ for $\tau(\mP)$ and $\lambda(\mP)$, respectively.
We shall also refer $1-\lambda(\mP)$ as the \emph{spectral gap} of $\mP$.

\section{Matrix Chernoff Bounds for Markov Chains}
\label{sec:chernoff}

This section provides a brief overview of our proof of Markov chain Martrix Chernoff bounds.
We start from a simpler version which only consider real-valued symmetric matrices,
as stated in Theorem~\ref{thm:chernoff} below. 
Then we extend it to complex-valued Hermitian matrices, as stated in  in Theorem~\ref{thm:complexchernoff}.

\begin{restatable}[A Real-Valued Version of Theorem~\ref{thm:complexchernoff}]{theorem}{chernoff}
    \label{thm:chernoff}
    Let $\mP$ be a regular Markov chain with state space  $[N]$,
    stationary distribution $\vpi$ and spectral expansion $\lambda$.
    Let $f: [N] \rightarrow \mathbb{R}^{d\times d}$
    be a function such that (1) $\forall v \in [N]$, $f(v)$ is symmetric and $\norm{f(v)}{2}\leq 1$; (2) $\sum_{v\in [N]} \pi_v f(v) = 0$.
    Let $(v_1,\cdots, v_k)$ denote a $k$-step random walk on $\mP$ starting from a distribution $\vphi$ on $[N]$.
    Then given $\eps\in (0, 1)$,
    \beq{
        \nonumber
        \besp{
            \mathbb{P}\left[\lambda_{\max}\left( \frac{1}{k}\sum_{j=1}^k f(v_j)\right)\geq \epsilon\right] &\leq \norm{\vphi}{\vpi} d^{2}\exp{\left( -(\eps^2 (1-\lambda)k / 72) \right)} \\
            \mathbb{P}\left[\lambda_{\min}\left( \frac{1}{k}\sum_{j=1}^k f(v_j)\right)\leq -\epsilon\right] &\leq \norm{\vphi}{\vpi}d^{2} \exp{\left( - (\eps^2 (1-\lambda)k / 72) \right)}.
        }}
\end{restatable}

Due to space constraints,
we defer the full proof to
\Secref{sec:chernoff_proof_all} in the supplementary material
and instead present a sketch here.
By symmetry, we only discuss
on bounding $\lambda_{\max}$ here.
Using the exponential method, the probability in Theorem~\ref{thm:chernoff} can be upper bounded for
any $t > 0$ by:
\begin{equation}
    \scriptsize
    \nonumber
    \besp{
        \mathbb{P}\left[\lambda_{\max}\left( \frac{1}{k}\sum_{j=1}^k f(v_j)\right)\geq \epsilon\right]
        \leq   \mathbb{P}\left[\Tr{\left[\exp{\left( t\sum_{j=1}^k f(v_j)\right)}\right]}\geq \exp{(tk\epsilon)}\right]
        \leq  \frac{\mathbb{E}\left[\Tr{\left[\exp{\left( t\sum_{j=1}^k f(v_j)\right)}\right]}\right]}{\exp{(tk\epsilon)}},
    }
    \normalsize
\end{equation}where the first inequality follows by the tail bounds for eigenvalues~(See Proposition 3.2.1 in \citet{tropp2015introduction})
which controls the tail probabilities of the extreme eigenvalues of a random matrix
by producing a bound for the trace of the matrix moment generating function,
and the second inequality follows by Markov's inequality. 
The RHS of the above equation is the expected trace 
of the exponential of a sum of matrices~(i.e., $tf(v_j)$'s).
When $f$ is a scalar-valued function, we can easily write  exponential of a sum to be product of exponentials~(since $\exp(a+b) = \exp(a)\exp(b)$ for scalars).
However, this is not true for matrices.
To bound the expectation term,
we invoke the following multi-matrix Golden-Thompson inequality from~\cite{garg2018matrix},
by letting $\mH_j=tf(v_j), j\in[k]$.
\begin{restatable}[Multi-matrix Golden-Thompson Inequality, Theorem~1.5 in \cite{garg2018matrix}]{theorem}{goldenthompson}
    \label{thm:golden_thompson}
    Let $\mH_1, \cdots \mH_k$ be $k$ Hermitian matrices, then for some probability distribution $\mu$  on $[-\frac{\pi}{2}, \frac{\pi}{2}]$.
    \beq{
        \nonumber
        \log{\left(\Tr{\left[\exp{\left(\sum_{j=1}^k \mH_j \right)}\right]}\right)} \leq \frac{4}{\pi} \int_{-\frac{\pi}{2}}^{\frac{\pi}{2}}
        \log{\left( \Tr{\left[ \prod_{j=1}^k \exp{\left( \frac{e^{\iu \phi}}{2} \mH_j\right)} \prod_{j=k}^1 \exp{\left( \frac{e^{-\iu \phi}}{2} \mH_j\right)}\right]}\right)} d\mu(\phi).
    }
\end{restatable}
The key point of this theorem is to relate the exponential
of a sum of matrices to a product of matrix exponentials and their adjoints,
whose trace can be further bounded via the following lemma
by letting $e^{\iu \phi} = \gamma + \iu b$.
\begin{restatable}[Analogous to Lemma 4.3 in \cite{garg2018matrix}]{lemma}{garglemma}
    \label{lemma:stoc18_lemma_4.3}
    Let $\mP$ be a regular Markov chain with state space $[N]$ with spectral expansion $\lambda$. Let $f$ be a function $f: [N] \rightarrow \R^{d\times d}$
    such that  (1) $\sum_{v\in [N]} \pi_v f(v)= 0$; (2) $\norm{f(v)}{2}\leq 1$ and $f(v)$ is symmetric, $v\in[N]$.
    Let $(v_1,\cdots, v_k)$ denote a $k$-step random walk on $\mP$ starting from a distribution $\vphi$ on $[N]$.
    Then for any $t>0, \gamma\geq 0, b>0$ such that
    $t^2(\gamma^2 + b^2)\leq 1$ and $t\sqrt{\gamma^2 + b^2}\leq \frac{1-\lambda}{4\lambda}$, we have
\scriptsize
\begin{equation}
        \nonumber
        \E
        \left[\Tr
            \left[
                \prod_{j=1}^k \exp{\left(\frac{tf(v_j)(\gamma + \iu b)}{2}\right)}
                \prod_{j=k}^1 \exp{\left(\frac{tf(v_j)(\gamma - \iu b)}{2}\right)}
                \right]
            \right]
        \leq \norm{\vphi}{\vpi}d \exp{\left(  kt^2(\gamma^2 + b^2)\left(1+\frac{8}{1-\lambda}\right)\right)}.
\end{equation}
\normalsize
\end{restatable}
Proving Lemma~\ref{lemma:stoc18_lemma_4.3} is
the technical core of our paper.
The main idea is to write the expected trace expression in LHS of Lemma~\ref{lemma:stoc18_lemma_4.3} in terms of the transition 
probability matrix $\mP$,
which allows for a recursive analysis 
to track how much the expected trace expression changes as a function of $k$.
The analysis relies on incorporating the
concentration of matrix-valued functions from~\cite{garg2018matrix} into the study
of general Markov chains from~\cite{chung2012chernoff},
which was originally for scalars.
Key to this extension is the definition of an inner product related to the stationary distribution $\vpi$ of $\mP$, 
and a spectral expansion from such inner products.
In contrast, the undirected regular graph case studied in \cite{garg2018matrix} can be handled using the standard
inner products, as well as the second largest eigenvalues
of $\mP$ instead of the spectral expansion.
Detailed proofs of Theorem~\ref{thm:chernoff} and Lemma~\ref{lemma:stoc18_lemma_4.3} 
can 
be found in Appendix~\ref{sec:proof_chernoff} and Appendix~\ref{sec:proof_lemma} of the supplementary material, respectively.

Our result about real-valued matrices can be further generalized to complex-valued matrices, as stated in Theorem~\ref{thm:complexchernoff}.
Our main strategy  is to adopt complexification technique~\cite{dongarra1984eigenvalue},
which first relate the eigenvalues of a $d\times d$ complex Hermitian matrix 
to a $2d \times 2d$ real symmetric matrix, and then deal with the real symmetric matrix
using Theorem~\ref{thm:chernoff}. 
The proof of Theorem~\ref{thm:complexchernoff} is deferred to Appendix~\ref{sec:proof_complex_chernoff} in the supplementary material.

\section{Convergence Rate of Co-occurrence Matrices of Markov Chains}
\label{sec:rate}

In this section, we first apply the matrix Chernoff bound
for regular Markov chains from Theorem~\ref{thm:chernoff}
to obtain our main result on the convergence of co-occurrence
matrix estimation, as stated in Theorem~\ref{thm:rate}, 
and then discuss its generalization to Hidden Markov models in Corollary~\ref{col:hmm}.
Informally, our result in Theorem~\ref{thm:chernoff} states that if the mixing time of the Markov chain $\mP$ is $\tau$,
then the length of a trajectory needed to guarantee an additive error~(in 2-norm) of $\epsilon$ is
 roughly $O\left((\tau+T)(\log{n} + \log{\tau +T})/\eps^2 + T\right)$, where $T$ is the co-occurrence window size.
%
However, we cannot directly apply the matrix Chernoff bound  because the co-occurrence matrix is not a sum of matrix-valued functions sampled from the original Markov chain $\mP$. The main difficulty is to construct the proper Markov chain and matrix-valued function as desired by Theorem~\ref{thm:chernoff}. We formally give our proof as follows:

\begin{proof}
(of Theorem~\ref{thm:rate})
    Our proof has three main steps:
    the first two construct
    a Markov chain $\mQ$ according to $\mP$, and 
    a matrix-valued function $f$ such that
    the sums of matrix-valued random variables sampled
    via $\mQ$ is exactly the error matrix
    $\mC - \mathbb{AE}[\mC]$.
    Then we invoke Theorem~\ref{thm:chernoff} to the
    constructed Markov chain $\mQ$ and 
    function $f$ to bound the convergence rate.
    We give details 
    below.

    \vpara{Step One}
    Given a random walk $(v_1, \cdots, v_L)$ on Markov chain $\mP$, we construct a sequence $(X_1, \cdots, X_{L-T})$ where
    $X_i \triangleq (v_i, v_{i+1}, \cdots, v_{i+T})$, i.e., each $X_i$ is a size-$T$ sliding window over  $(v_1, \cdots, v_L)$.
    Meanwhile, let $\mathcal{S}$ be the set of all $T$-step walks on Markov chain $\mP$, we define a new Markov chain $\mQ$ on $\mathcal{S}$
    such that $\forall (u_0, \cdots, u_T), (w_0, \cdots, w_T) \in \mathcal{S}$:
    \beq{
        \nonumber
        \mQ_{(u_0, \cdots, u_T), (w_0, \cdots, w_T)} \triangleq \begin{cases}
            \mP_{u_T, w_T} & \text{if } (u_1, \cdots, u_T) = (w_0, \cdots, w_{T-1}); \\
            0              & \text{otherwise}.
        \end{cases}
    }The following claim characterizes the properties of $\mQ$, whose proof is deferred to Appendix~\ref{subsec:proof_claimQ} in the supplementary material.
    \begin{restatable}[Properties of $\mQ$]{claim}{propertyQ}
        \label{claim:Q}
        If $\mP$ is a regular Markov chain,  then $\mQ$ satisfies:
        \begin{enumerate}[leftmargin=*,itemsep=0pt,parsep=0.5em,topsep=0.3em,partopsep=0.3em]
            \item  $\mQ$ is a regular Markov chain with stationary distribution 
            $\sigma_{(u_0, \cdots, u_T)} = \pi_{u_0} \mP_{u_0, u_1}\cdots \mP_{u_{T-1}, u_T}$;
            \item The sequence $(X_1, \cdots X_{L-T})$ is a random walk on $\mQ$ starting from a distribution $\vrho$ such that
            $\rho_{(u_0, \cdots, u_T)} = \phi_{u_0} \mP_{u_0, u_1}\cdots \mP_{u_{T-1}, u_T}$, and $\norm{\vrho}{\vsigma}= \norm{\vphi}{\vpi}$.
            \item $\forall \delta > 0$, the $\delta$-mixing time of $\mP$ and $\mQ$ satisfies $\tau(\mQ) < \tau(\mP)+T$;
            \item $\exists \mP$ with $\lambda(\mP)<1$ s.t. the induced $\mQ$ has $\lambda(\mQ)=1$, i.e. $\mQ$ may 
            have zero spectral gap.
        \end{enumerate}
    \end{restatable}
    Parts $1$ and $2$ 
    imply that the sliding windows~(i.e., $X_1, X_2, \cdots$) correspond to the state transition in a regular Markov chain $\mQ$, whose 
     mixing time and
    spectral expansion are described in Parts $3$ and $4$. 
    A special case of the above construction when $T=1$ can be found in Lemma 6.1 of \cite{wolfer2019estimating}.

    \vpara{Step Two}
    Defining a matrix-valued function $f: \mathcal{S} \rightarrow \R^{n\times n}$ such that  $\forall X=(u_0, \cdots, u_T) \in \mathcal{S}$:
    \beq{
        \label{eq:matrix_value_func}
        f(X) \triangleq \frac{1}{2}\left(\sum_{r=1}^T\frac{\alpha_r}{2}\left(\ve_{u_0} \ve^\top_{u_{r}} + \ve_{u_{r}} \ve^\top_{u_0}\right) - \sum_{r=1}^T\frac{\alpha_r}{2}  \left(\mPi \mP^r + \left(\mPi\mP^r\right)^\top \right)\right).
    }With this definition of $f(X)$, the difference between
    the co-occurrence matrix $\mC$ and its asymptotic expectation $\mathbb{AE}[\mC]$ can be written as:
    $\mC - \mathbb{AE}[\mC]
    =2(\frac{1}{L-T}\sum_{i=1}^{L-T} f(X_i))$.
    We can further show the following properties of this
    function $f$:
    \begin{restatable}[Properties of $f$]{claim}{propertyf}
        \label{claim:f}
        The function $f$ in \Eqref{eq:matrix_value_func} satisfies (1) $\sum_{X\in \mathcal{S}} \sigma_X f(X) = 0$;
        (2) $f(X)$ is symmetric and $\norm{f(X)}{2} \leq 1, \forall X\in \mathcal{S}$.
    \end{restatable}
    This claim verifies that $f$  in \Eqref{eq:matrix_value_func} 
    satisfies the two conditions of matrix-valued function in Theorem~\ref{thm:chernoff}.
    The proof of Claim~\ref{claim:f} is deferred to
    Appendix~\ref{subsec:proof_claimf}  of the supplementary material.

    \vpara{Step Three} The construction in step two reveals the fact that the error matrix $\mC -\mathbb{AE}[\mC]$
    can be written as the average of matrix-valued random variables~(i.e., $f(X_i)$'s), which are sampled via a regular Markov chain $\mQ$ 
    This  encourages us
    to directly apply Theorem~\ref{thm:chernoff}.
    However, note that (1) the error probability in Theorem~\ref{thm:chernoff} contains a factor
    of spectral gap $(1 - \lambda)$; and (2)
    Part 4 of Claim~\ref{claim:Q} allows for the existence
    of a Markov chain $\mP$ with $\lambda(\mP) < 1$ while the induced Markov chain $\mQ$ has $\lambda(\mQ)=1$. So we cannot directly apply Theorem~\ref{thm:chernoff} to \(\mQ\).
    To address this issue, we utilize the following
    tighter bound on sub-chains.
    \begin{claim}(Claim 3.1 in \citet{chung2012chernoff})
        Let $\mQ$ be a regular Markov chain with $\delta$-mixing time $\tau(Q)$, then $\lambda\left(\mQ^{\tau(Q)}\right) \leq \sqrt{2\delta}$.
        In particular,
        setting $\delta=\frac{1}{8}$ implies $\lambda(\mQ^{\tau(\mQ)})\leq \frac{1}{2}$.
    \end{claim}
    The above claim reveals the fact that, even though $\mQ$ could have zero spectral gap~(Part 4 of Claim~\ref{claim:Q}),
    we can bound the spectral expansion of $\mQ^{\tau(\mQ)}$.
    We partition $(X_1, \cdots X_{L-T})$ into $\tau(\mQ)$ groups\footnote{Without loss of generality, we assume $L-T$ is a multiple of $\tau(\mQ)$.},
    such that the $i$-th group consists of a sub-chain $(X_i, X_{i+\tau(\mQ)}, X_{i+2\tau(Q)}, \cdots)$ of length $k\triangleq(L-T)/\tau(\mQ)$.
    The sub-chain can be viewed as generated from a Markov chain $\mQ^{\tau(Q)}$.
    Apply Theorem~\ref{thm:chernoff} to the $i$-th sub-chain,
    whose starting distribution is $\vrho_i \triangleq \left(\mQ^\top\right)^{i-1} \vrho$, we have
    \beq{
        \nonumber
        \besp{
        \mathbb{P}\left[\lambda_{\max}\left( \frac{1}{k}\sum_{j=1}^{k} f(X_{i+(j-1)\tau(\mQ)}\right) \geq \eps\right]
        &\leq \norm{\vrho_i}{\vsigma} n^{2} \exp{\left( -\eps^2 \left(1-\lambda\left(\mQ^{\tau(\mQ)}\right)\right)k/72 \right)}\\
        &\leq \norm{\vrho_i}{\vsigma} n^{2}\exp{\left( -\eps^2k/144 \right)}\leq \norm{\vphi}{\vpi} n^{2}\exp{\left( -\eps^2k/144 \right)},
    }}where that last step follows by $\norm{\vrho_i}{\vsigma} \leq \norm{\vrho_{i-1}}{\vsigma} \leq \cdots \norm{\vrho_1}{\vsigma} = \norm{\vrho}{\vsigma}$ 
    and $\norm{\vrho}{\vsigma}= \norm{\vphi}{\vpi}$~(Part 2 of Claim~\ref{claim:Q}). Together with a union bound across each sub-chain, we can obtain:
    \beq{\nonumber
        \besp{
            &\mathbb{P}\left[\lambda_{\max}\left( \mC- \mathbb{AE}[\mC]\right)\geq \epsilon\right]  
            =\mathbb{P}\left[\lambda_{\max}\left( \frac{1}{L-T}\sum_{j=1}^{L-T} f(X_j)\right)\geq \frac{\eps}{2}\right] \\
            = &\mathbb{P}\left[\lambda_{\max}\left(\frac{1}{\tau(\mQ)}\sum_{i=1}^{\tau(\mQ)} \frac{1}{k}\sum_{j=1}^{k} f(X_{i+(j-1)\tau(\mQ)})\right)\geq \frac{\eps}{2}\right]\\
            \leq &\sum_{i=1}^{\tau(\mQ)} \mathbb{P}\left[\lambda_{\max}\left( \frac{1}{k}\sum_{j=1}^{k} f(X_{i+(j-1)N})\right)\geq \frac{\eps}{2}\right]  \leq \tau(\mQ) \norm{\vphi}{\vpi} n^{2} \exp{\left( -\eps^2k/576 \right)}.
        }}The bound on $\lambda_{\min}$ also follows similarly.
    As $\mC-\mathbb{AE}[\mC]$ is a real symmetric matrix,
    its $2$-norm is its maximum absolute eigenvalue.
    Therefore, we can use the eigenvalue bound to bound
    the overall error probability in terms of the matrix 2-norm:
    \beq{\nonumber
        \besp{
        &\mathbb{P}\left[\norm{\mC-\mathbb{AE}[\mC]}{2} \geq \eps\right] 
        = \mathbb{P}\left[\lambda_{\max}(\mC-\mathbb{AE}[\mC]) \geq \eps \lor  \lambda_{\min}(\mC-\mathbb{AE}[\mC]) \leq -\eps\right]\\
        \leq& 2\tau(\mQ) n^{2} \norm{\vphi}{\vpi}\exp{\left( -\eps^2k/576 \right)}
        \leq 2\left(\tau(\mP) + T\right)\norm{\vphi}{\vpi} n^{2} \exp{\left( -\frac{\eps^2(L-T)}{576\left(\tau(\mP) + T\right)} \right)},
    }}where the first inequality follows by union bound,
    and the second inequality is due to $\tau(\mQ) < \tau(\mP) + T$~(Part 3 of Claim~\ref{claim:Q}).
    This bound implies that the probability that $\mC$ deviates  from $\mathbb{AE}[\mC]$ could be arbitrarily
    small by increasing the sampled trajectory length $L$. 
    Specially, if we want the event $\norm{\mC-\mathbb{AE}[\mC]}{2} \geq \eps$ 
    happens with probability smaller than $1/n^{O(1)}$, we need
    $
        L = O\left(\left(\tau(\mP)+T\right)\left(\log{n} + \log{\left(\tau(\mP)+T\right)}\right)/\eps^{2}+ T\right)
    $. If we assume $T=O(1)$, we can achieve $L = O\left(\tau(\mP)\left(\log{n} + \log{\tau(\mP)}\right)/\eps^{2}\right)$.
\end{proof} 
Our analysis can be extended to Hidden Markov models~(HMM) as shown in Corollary~\ref{col:hmm}, and has a potential to solve problems raised in \cite{huang2018learning, mattila2020fast}. Our strategy is to treat the HMM with observable state space $\mathcal{Y}$ and hidden state space $\mathcal{X}$ as a Markov chain with state space $\mathcal{Y}\times \mathcal{X}$. The detailed proof can be found in Appendix~\ref{subsec:proof_hmm} in the supplementary material.
\begin{restatable}[Co-occurrence Matrices of HMMs]{corollary}{hmm}
\label{col:hmm}
For a HMM with observable states $y_t\in \mathcal{Y}$ and hidden states $x_t\in\mathcal{X}$,
let $P(y_t|x_t)$ be the emission probability and $P(x_{t+1}|x_{t})$ be the hidden state transition probability.
Given an $L$-step trajectory observations from the HMM, $(y_1, \cdots, y_L)$, one needs a trajectory of length $L=O(\tau (\log{|\mathcal{Y}|} + \log{\tau}) / \epsilon^2)$ to achieve a co-occurrence matrix  within error bound $\epsilon$ with high probability,
where $\tau$ is the mixing time of the Markov chain on hidden states.
\end{restatable}

\section{Experiments}
\label{sec:exp}

\begin{figure}[t]
    \centering
    \subfloat[Barbell Graph]{
        \centering
        \label{fig:barbel}
        \includegraphics[width=.24\textwidth]{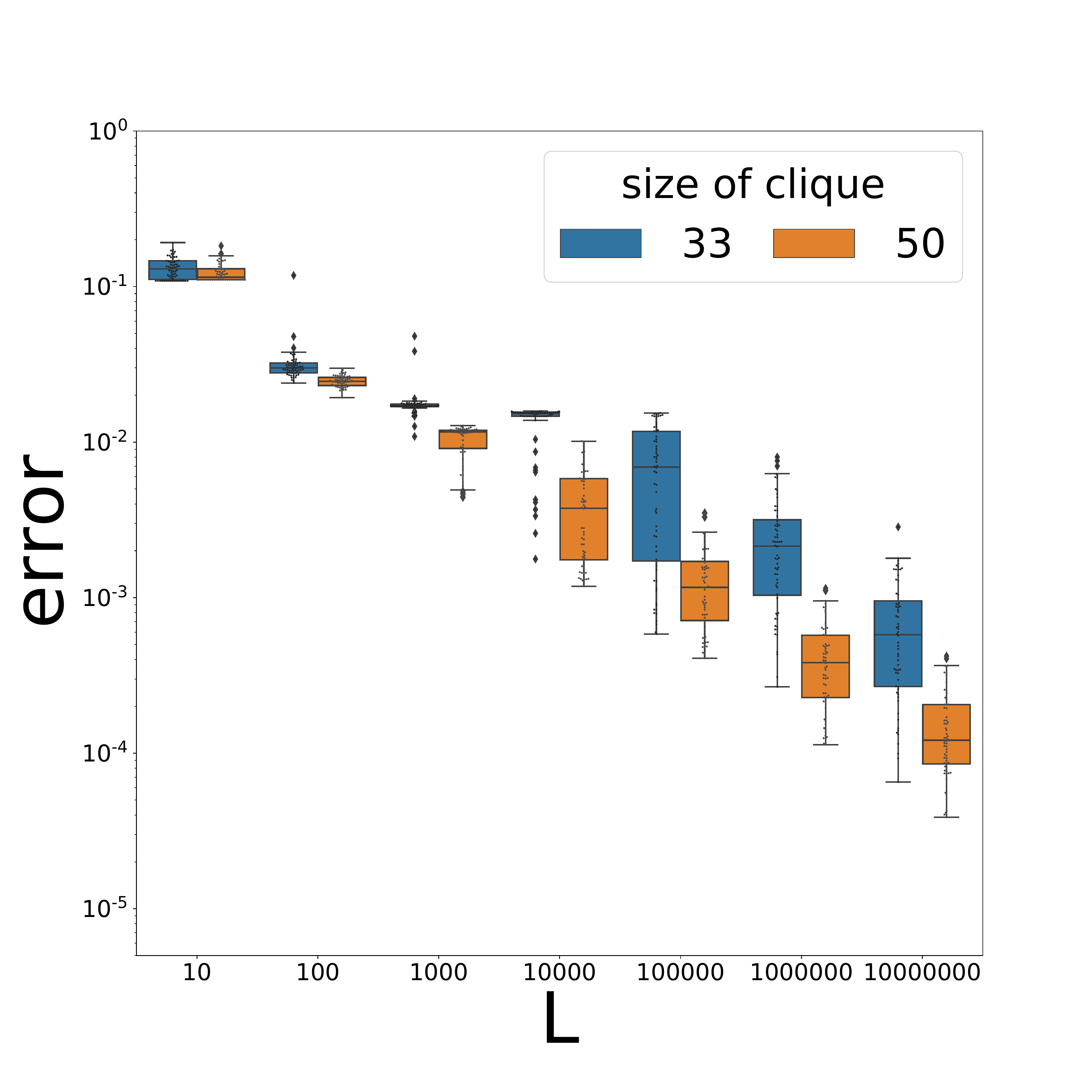}
    }    
    \subfloat[Winning Streak Chain]{
        \label{fig:winning}
        \includegraphics[width=.24\textwidth]{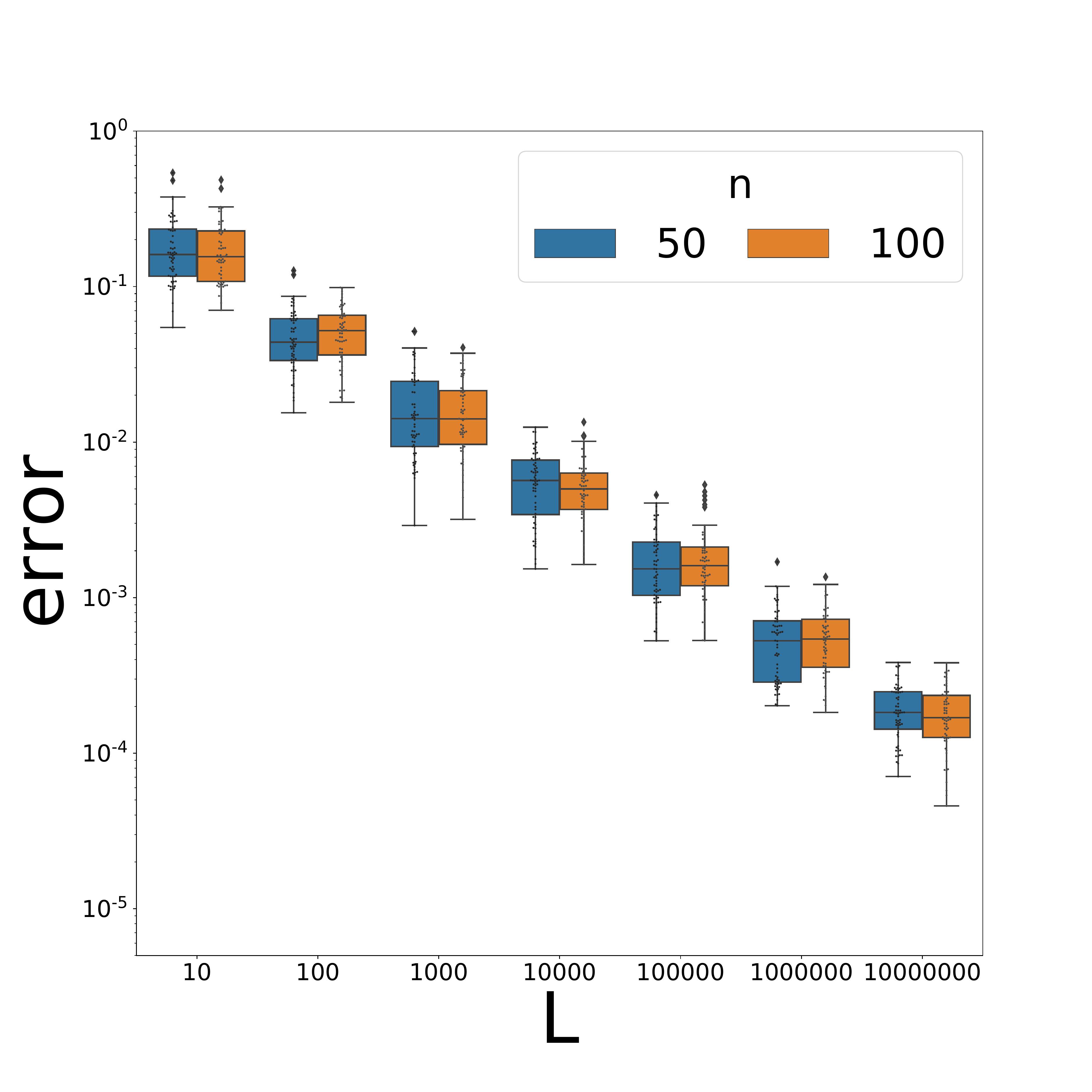}
    }
    \subfloat[BlogCatalog]{
        \label{fig:blog}
        \includegraphics[width=.24\textwidth]{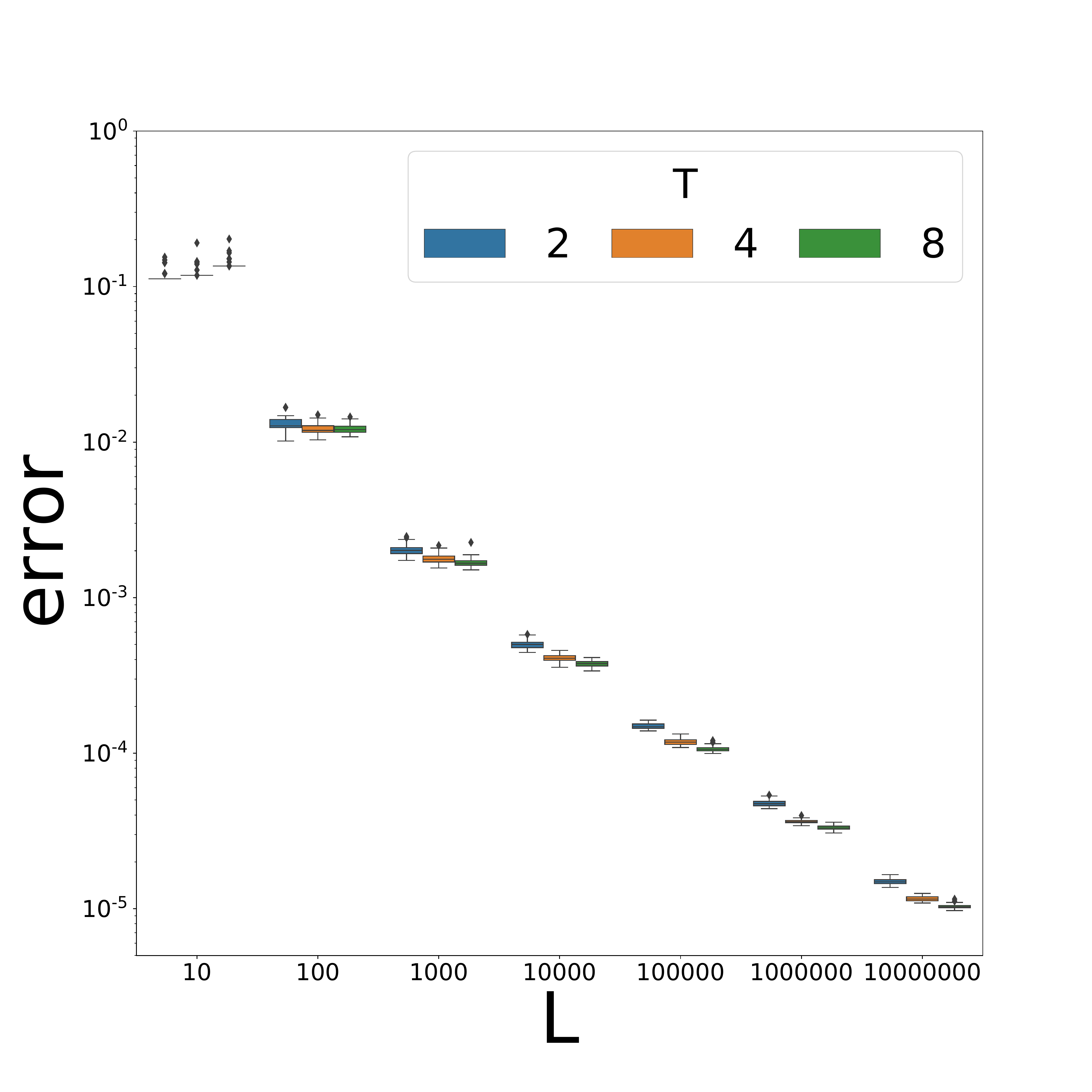}
    }
    \subfloat[Random Graph]{
        \label{fig:gnp}
        \includegraphics[width=.24\textwidth]{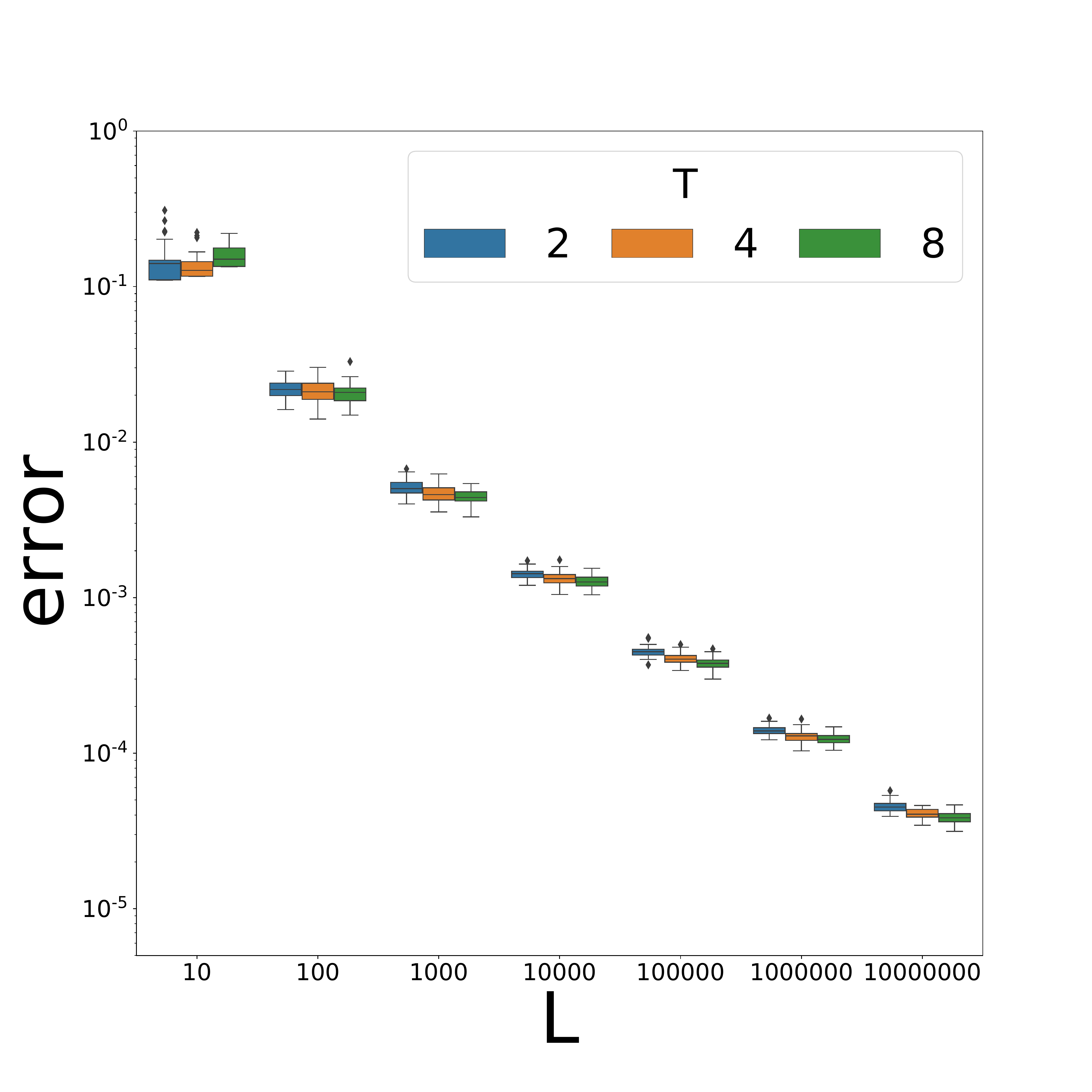}
    }
    \caption{The convergence rate of co-occurrence matrices on Barbell graph, winning streak chain, BlogCatalog graph
    , and random graph~(in log-log scale). 
    The $x$-axis is the trajectory length $L$ and the $y$-axis is the  error $\norm{\mC-\mathbb{AE}[\mC]}{2}$. Each experiment contains 64 trials, and the error bar is presented.}
    \label{fig:exp}
\end{figure}

In this section, we show experiments to illustrate the exponentially fast convergence rate 
of estimating co-occurrence matrices of Markov chains. 
We conduct experiments on three synthetic Markov chains~(Barbell graph, winning streak chain, and random graph) and one real-world Markov chain~(BlogCatalog). 
For each Markov chain and each trajectory length
$L$ from the set $\{10^1, \cdots, 10^7\}$,
we measure the approximation error of the 
co-occurrence matrix constructed by Algorithm~\ref{alg:dw}
from a $L$-step random walk sampled from the chain.
We performed 64 trials for each experiment,
and the results are aggregated as an  error-bar plot.
We set $T=2$ and $\alpha_r$ to be uniform unless otherwise mentioned.
The relationship between trajectory length $L$ and approximation error $\norm{\mC-\mathbb{AE}[\mC]}{2}$ is shown in~\Figref{fig:exp}~(in log-log scale).
Across all the four datasets,  the observed
exponentially fast convergence rates match what our bounds predict in Theorem~\ref{thm:rate}.
Below we discuss our observations
for each of these datasets.

\vpara{Barbell Graphs~\cite{sauerwald2019random}}
The Barbell graph is an undirected graph with two cliques connected by a single path.
Such graphs' mixing times vary greatly:
two cliques with size $k$ connected by a single edge have
mixing time $\Theta(k^2)$;
and two size-$k$ cliques connected by a length-$k$ path
have mixing time about $\Theta(k^3)$.
We evaluate the convergence rate of co-occurrence matrices
on the two graphs mentioned above,
each with $100$ vertices.
According to our bound that require $L=O(\tau(\log{n}+\log{\tau})/\eps^2)$, we shall expect
the approximate co-occurrence matrix to converge faster when
the path bridging the two cliques is shorter.
The experimental results are shown in \Figref{fig:barbel},
and indeed display faster convergences when the 
path is shorter~(since we fix $n=100$, a Barbell graph with clique size 50 has a shorter path 
connecting the two cliques than the one with clique size 33).

\vpara{Winning Streak Chains~(Section 4.6 of \cite{levin2017markov})}
A winning streak Markov chain has state space $[n]$,
and can be viewed as tracking the number of consecutive
`tails' in a sequence of coin flips.
Each state transits back to state $1$ with probability $0.5$,
and the next state with probability $0.5$.
The $\delta$-mixing time of this chain satisfies
$\tau \leq \lceil \log_2(1/\delta)\rceil$,
and is independent of $n$.
This prompted us to choose this chain, as we should expect
similar rates of convergence for different values of $n$
according to our bound of $L=O(\tau(\log{n}+\log{\tau})/\eps^2)$.
In our experiment, we compare between $n=50$ and $n = 100$
and illustrate the results in \Figref{fig:winning}.
As we can see, for each trajectory length $L$, the approximation errors of $n=50$ and $n=100$ are indeed very close.

\vpara{BlogCatalog Graph~\cite{tang2009relational}}
is widely used to benchmark graph representation learning algorithms~\cite{perozzi2014deepwalk,grover2016node2vec,qiu2018network}.
It is an undirected graph of social relationships of online
bloggers with 10,312 vertices and 333,983 edges.
The random walk on the BlogCatalog graph has spectral expansion $\lambda\approx 0.57$.
Following \citet{levin2017markov},
we can upper bound its $\frac{1}{8}$-mixing time by $\tau\leq 36$. 
We choose $T$ from $\{2, 4, 8\}$ and illustrate the results in \Figref{fig:blog}.
The convergence rate is robust to different values of $T$. 
Moreover, the variance in BlogCatalog is much smaller than that in other datasets.

We further demonstrate how our result could be used to select parameters for a popular graph representation learning algorithm, DeepWalk~\cite{perozzi2014deepwalk}. 
We set the window size $T=10$, which is the default value of DeepWalk.
Our bound on trajectory length $L$ in Theorem 1~(with explicit constant) is $L \geq 576(\tau + T) (3\log{n} + \log{(\tau+T)})/\epsilon^2 + T$.
The error bound $\epsilon$ might be chosen in the range of $[0.1,0.01]$, which corresponds to $L$ in the range of $[8.4\times 10^7, 8.4\times 10^9]$. 
To verify that is a meaningful range for tuning $L$, we enumerate trajectory length $L$ from $\{10^4, \cdots, 10^{10}\}$, estimate the co-occurrence matrix with the single trajectory sampled from BlogCatalog, convert the co-occurrence matrix to the one implicitly factorized by DeepWalk~\cite{perozzi2014deepwalk, qiu2018network}, and factorize it with SVD. For comparison, we also provide the result at the limiting case~($L\rightarrow +\infty$) where we directly compute the asymptotic expectation of the co-occurrence matrix according to \Eqref{eq:expected}. The limiting case involves computing a matrix polynomial and could be very expensive.
For node classification task, the micro-F1 when training ratio is 50\% is
\begin{center}
\small
\begin{tabular}{c|ccccccc|c} 
 \toprule
 Length $L$ of DeepWalk & $10^4$ & $10^5$ & $10^6$ & $10^7$ & $10^8$ & $10^9$ & $10^{10}$ & $+\infty$ \\ 
 Micro-F1~($\%$) & 15.21 & 18.31 & 26.99  & 33.85 & 39.12 & 41.28 & 41.58 & \textbf{41.82} \\
 \bottomrule
\end{tabular}.
\normalsize
\end{center}
As we can see, it is reasonable to choose $L$ in the predicted range. 

\vpara{Random Graph}
The small variance observed on BlogCatalog
leads us to hypothesize that it shares some traits
with random graphs.
To gather further evidence for this, we estimate the
co-occurrence matrices of an Erdős–Rényi
random graph for comparison.
Specifically, we take a random graph on $100$ vertices
where each undirected edge is added independently with probability $0.1$, aka. $G(100, 0.1)$.
The results~\Figref{fig:gnp} show very similar
behaviors compared to the BlogCatalog graph:
small variance and robust convergence rates.

\section{Conclusion and Future Work}
\label{sec:conclusion}

In this paper, we analyze the convergence rate of estimating the co-occurrence matrix
of a regular Markov chain.
The main technical contribution of our work is to prove  a
Chernoff-type bound for sums of matrix-valued random variables sampled via a regular Markov chain,
and we show that the problem of estimating co-occurrence matrices
is a non-trivial application of the Chernoff-type bound.
Our results show that, given a regular Markov chain with $n$ states and mixing time $\tau$, 
we need a trajectory of length $O(\tau (\log{n} + \log{\tau})/\eps^2)$ to achieve an estimator of the co-occurrence matrix with error bound $\epsilon$.
Our work leads to some natural future questions:
\begin{itemize}[leftmargin=*,itemsep=0pt,parsep=0.5em,topsep=0.3em,partopsep=0.3em]
\item Is it a tight bound? Our analysis on convergence rate of co-occurrence matrices relies on union bound, which probably gives a loose bound.
 It would be interesting to shave off the leading factor $\tau$ in the bound, as the mixing time $\tau$ could be large for some Markov chains.
\item What if the construction of the co-occurrence matrix is coupled with a learning algorithm? For example, in word2vec~\cite{NIPS2013_5021}, the co-occurrence in each sliding window outputs a mini-batch to a logistic matrix factorization model. This problem can be formalized as the convergence of stochastic gradient descent with non-i.i.d but Markovian random samples.
\item Can we find more applications of the Markov chain matrix Chernoff bound?  
We believe Theorem~\ref{thm:chernoff} could have further applications, e.g., in reinforcement learning~\cite{ortner2020regret}. 
\end{itemize}

\section*{Broader Impact}

Our work contributes to the research literature of Chernoff-type bounds and co-occurrence statistics.
Chernoff-type bound have become one of the most important probabilistic results in computer science. Our result generalize Chernoff bound to Markov dependence and random matrices.
Co-occurrence statistics have emerged as important tools in machine learning.
Our work  addresses the sample complexity of estimating co-occurrence matrix.
We believe such better theoretical understanding can
further the understanding of potential and limitations of
graph representation learning and reinforcement learning.

\comment{
Graph representation learning is one of the most widely used tools for analyzing large networks.
The theoretical behaviors of many of these tools are not well
understood at the moment, and as a result, many existing
network analytic tools fall outside of provable envelopes.
Our work addresses the sample complexity of one of the first
such algorithms, DeepWalk.
We believe such better theoretical understanding can
further the understanding of potential and limitations of
graph representation learning.
}
\begin{ack}
We thank Jian Li~(IIIS, Tsinghua) and 
Shengyu Zhang~(Tencent Quantum Lab) for motivating this work.
Funding in direct support of this work:
Jiezhong Qiu and Jie Tang were supported by the
National Key R\&D Program of China (2018YFB1402600),
NSFC for Distinguished Young Scholar 
(61825602),
and NSFC (61836013).
Richard Peng was partially supported by NSF grant CCF-1846218.
There is no additional revenue related to this work.
\end{ack}
\newpage
\bibliographystyle{plainnat}
\bibliography{ref}

\newpage

\appendix
\section*{Supplementary Material of A Matrix Chernoff Bound for Markov Chains and Its
Application to Co-occurrence Matrices}

\section{Convergence Rate of Co-occurrence Matrices}
\label{sec:proof_cooccurrence}

\subsection{Proof of Claim~\ref{claim:Q}}
\label{subsec:proof_claimQ}
\propertyQ*
\begin{proof} We prove the fours parts of this Claim one by one.

    \vpara{Part 1}
    To prove $\mQ$ is regular, it is sufficient to show that $\exists N_1$, $\forall n_1>N_1$, $(v_0, \cdots, v_T)$ can reach $(u_0, \cdots, u_T)$ at $n_1$ steps.
    We know $\mP$ is a regular Markov chain,
    so there exists $N_2 \geq T$ s.t., for any $n_2\geq N_2$, $v_T$ can reach $u_0$ at exact $n_2$ step, i,e.,
    there is a $n_2$-step walk s.t. $(v_T, w_1, \cdots, w_{n_2-1}, u_0)$ on $\mP$. This induces an $n_2$-step walk
    from $(v_0, \cdots, v_T)$ to $(w_{n_2-T+1}, \cdots, w_{n_2-1}, u_0)$. Take further $T$ step,  we can reach $(u_0, \cdots, u_T)$,
    so we construct a $n_1=n_2+T$ step walk from $(v_0, \cdots, v_T)$ to $(u_0, \cdots u_T)$.
    Since this is true for any $n_2 \geq N_2$,
    we then claim that any state can be reached from any other state in any number of
    steps greater than or equal to a number $N_1=N_2+T$.
    Next to verify $\vsigma$ such that $\sigma_{(u_0, \cdots, u_T)} = \pi_{u_0} \mP_{u_0, u_1}\cdots \mP_{u_{T-1}, u_T}$ is the stationary distribution of Markov chain $\mQ$,
    \beq{
        \nonumber
        \besp{
            &\sum_{(u_0, \cdots, u_T)\in \mathcal{S}}\sigma_{(u_0, \cdots, u_T)} \mQ_{(u_0, \cdots, u_T), (w_0, \cdots, w_T)}  \\
            =& \sum_{u_0: (u_0, w_0, \cdots, w_{T-1})\in \mathcal{S}} \pi_{u_0} \mP_{u_0, w_0} \mP_{w_0, w_1}, \cdots, \mP_{w_{T-2}, w_{T-1}}  \mP_{w_{T-1}, w_T}\\
            =& \left(\sum_{u_0} \pi_{u_0} \mP_{u_0, w_0}\right) \mP_{w_0, w_1}, \cdots, \mP_{w_{T-2}, w_{T-1}} \mP_{w_{T-1}, w_T}\\
            =& \pi_{w_0}\mP_{w_0, w_1}, \cdots, \mP_{w_{T-2}, w_{T-1}} \mP_{w_{T-1}, w_T} = \sigma_{w_0, \cdots, w_T}.
        }}

    \vpara{Part 2}  Recall $(v_1, \cdots, v_L)$ is a random walk on $\mP$ starting from distribution $\vphi$, so the probability we observe $X_1 = (v_1, \cdots, v_{T+1})$ is
    $\phi_{v_1} \mP_{v_1, v_2} \cdots \mP_{v_{T}, v_T} = \rho_{(v_1, \cdots, v_{T+1})}$, i.e., $X_1$ is sampled from the  distribution $\vrho$.
    Then we study the transition probability from $X_i=(v_i, \cdots, v_{i+T})$ to $X_{i+1}=(v_{i+1}, \cdots, v_{i+T+1})$, which is
    $\mP_{v_{i+T}, v_{i+T+1}} = \mQ_{X_i, X_{i+1}}$. Consequently, we can claim $(X_i, \cdots, X_{L-T})$ is a random walk on $\mQ$. Moreover,
    \beq{\nonumber
    \besp{
    \norm{\vrho}{\vsigma}^2 &= \sum_{(u_0, \cdots, u_T)\in \mathcal{S}} \frac{\rho^2_{(u_0, \cdots, u_T)}}{\sigma_{(u_0, \cdots, u_T)}} =\sum_{(u_0, \cdots, u_T)\in \mathcal{S}} \frac{\left(\phi_{u_0} \mP_{u_0, u_1} \cdots \mP_{u_{T-1}, u_T}\right)^2}{\pi_{u_0} \mP_{u_0, u_1} \cdots \mP_{u_{T-1}, u_T} } \\
    &= \sum_{u_0} \frac{\phi^2_{u_0}}{\pi_{u_0}} \sum_{(u_0, u_1,\cdots, u_T) \in \mathcal{S}} \mP_{u_0, u_1} \cdots \mP_{u_{T-1}, u_T} = \sum_{u_0} \frac{\phi^2_{u_0}}{\pi_{u_0}}  = \norm{\vphi}{\vpi}^2,
    }}which implies $\norm{\vrho}{\vsigma}= \norm{\vphi}{\vpi}$.

    \vpara{Part 3}
    For any distribution $\vy$ on $\mathcal{S}$,
    define  $\vx \in \R^n$ such that
    $x_i = \sum_{(v_1, \cdots, v_{T-1}, i) \in \mathcal{S}} y_{v_1, \cdots, v_{T-1}, i}$.
    Easy to see $\vx$ is a probability vector, since $\vx$ is the marginal probability of $\vy$.
    For convenience, we assume for a moment  the $\vx, \vy, \vsigma, \vpi$ are row vectors.
    We can see that:
    \beq{
        \nonumber
        \besp{
            \norm{\vy \mQ^{\tau(\mP)+T-1} - \vsigma}{TV} &= \frac{1}{2} \norm{\vy\mQ^{\tau(\mP)+T-1} - \vsigma}{1} \\
            &=\frac{1}{2}\sum_{(v_1, \cdots, v_T) \in \mathcal{S}} \abs{\left(\vy\mQ^{\tau(\mP)+T-1} - \vsigma\right)_{v_1, \cdots, v_T}}\\
            &=\frac{1}{2}\sum_{(v_1, \cdots, v_T) \in \mathcal{S}} \abs{\left(\vx\mP^{\tau(\mP)}\right)_{v_1}\mP_{v_1, v_2}\cdots \mP_{v_{T-1}, v_T} - \vpi_{v_1} \mP_{v_1, v_2}\cdots \mP_{v_{T-1}, v_T}}\\
            &=\frac{1}{2}\sum_{(v_1, \cdots, v_T) \in \mathcal{S}} \abs{\left(\vx\mP^{\tau(\mP)}\right)_{v_1}-\pi_{v_1}}\mP_{v_1, v_2}\cdots \mP_{v_{T-1}, v_T}\\
            &=\frac{1}{2}\sum_{v_1}\abs{\left(\vx\mP^{\tau(\mP)}\right)_{v_1}-\pi_{v_1}} \sum_{(v_1, \cdots, v_T) \in \mathcal{S}} \mP_{v_1, v_2}\cdots \mP_{v_{T-1}, v_T}\\
            &=\frac{1}{2}\sum_{v_1}\abs{\left(\vx\mP^{\tau(\mP)}\right)_{v_1}-\pi_{v_1}}=\frac{1}{2}\norm{\vx\mP^{\tau(\mP)} - \vpi}{1} = \norm{\vx\mP^{\tau(\mP)} - \vpi}{TV} \leq \delta.
        }
    }which indicates $\tau(\mQ) \leq \tau(\mP) + T - 1 < \tau(\mP) + T$.

    \vpara{Part 4}
    This is an example showing that $\lambda{(\mQ)}$ cannot be bounded by $\lambda{(\mP)}$ ---  even though $\mP$ has $\lambda{(\mP)} < 1$,
    the induced $\mQ$ may have $\lambda{(\mQ)}=1$. We consider random walk on the unweighted undirected graph $\fourEEzz$ and $T=1$.
    The transition probability matrix $\mP$ is:
    \makeatletter
    \renewcommand*\env@matrix[1][c]{\hskip -\arraycolsep
        \let\@ifnextchar\new@ifnextchar
        \array{*\c@MaxMatrixCols #1}}
    \makeatother
    \beq{
        \nonumber
        \mP = \begin{bmatrix}[r]
            0   & 1/3 & 1/3 & 1/3 \\
            1/2 & 0   & 1/2 & 0   \\
            1/3 & 1/3 & 0   & 1/3 \\
            1/2 & 0   & 1/2 & 0
        \end{bmatrix}
    }with stationary distribution $\vpi = \begin{bmatrix} 0.3 & 0.2 & 0.3 & 0.2\end{bmatrix}^\top$ and $\lambda(\mP)=\frac{2}{3}$.
    When $T=1$, the induced Markov chain $\mQ$ has stationary distribution
    $\sigma_{u, v} = \pi_u \mP_{u, v} = \frac{d_u}{2m} \frac{1}{d_u} = \frac{1}{2m}$ where $m=5$ is the number of edges in the graph.
    Construct $\vy \in \mathbb{R}^{\abs{\mathcal{S}}}$ such that
    \beq{
        \nonumber
        y_{(u, v)} = \begin{cases}
            1  & (u, v) = (0, 1),  \\
            -1 & (u, v) = (0, 3),  \\
            0  & \text{otherwise.}
        \end{cases}
    } The constructed vector $\vy$ has norm
    \beq{
        \nonumber
        \norm{\vy}{\vsigma} = \sqrt{\langle \vy, \vy\rangle_{\vsigma} } = \sqrt{\sum_{(u, v) \in \mathcal{S}} \frac{y_{(u, v)} y_{(u, v)} }{\sigma_{(u, v)}} } = \sqrt{\frac{ y_{(0, 1)} y_{(0, 1)} }{\sigma_{(0, 1)}} + \frac{ y_{(0, 3)} y_{(0, 3)} }{\sigma_{(0, 3)}}} = 2\sqrt{m}.
    } And it is easy to check $\vy \perp \vsigma$,
    since $\langle \vy, \vsigma \rangle_{\vsigma} = \sum_{(u, v) \in \mathcal{S}} \frac{\sigma_{(u, v)} y_{(u, v)} }{\sigma_{(u, v)}} = y_{(0, 1)}+y_{(0, 3)}=0$.
    Let $\vx =  \left(\vy^\ast \mQ\right)^\ast$, we have for $(u, v)\in \mathcal{S}$:
    \beq{
        \nonumber
        \vx_{(u, v)} = \begin{cases}
            1  & (u,v) = (1,2),    \\
            -1 & (u, v) = (3,2),   \\
            0  & \text{otherwise.}
        \end{cases}
    } This vector has norm:
    \beq{
        \nonumber
        \norm{ \vx}{\vsigma} = \sqrt{\langle  \vx, \vx \rangle_{\vsigma} } = \sqrt{\sum_{(u, v) \in \mathcal{S}} \frac{x_{(u, v)} x_{(u, v)} }{\sigma_{(u, v)}} } = \sqrt{\frac{ y_{(1, 2)} y_{(1, 2)} }{\sigma_{(1, 2)}} + \frac{ y_{(3, 2)} y_{(3, 2)} }{\sigma_{(3, 2)}}} = 2\sqrt{m}
    }Thus we have
    $\frac{\norm{\left(\vy^\ast \mQ\right)^\ast}{\vsigma}}{\norm{\vy}{\vsigma}} = 1$. Taking maximum over all possible $\vy$ gives $\lambda(\mQ) \geq 1$. Also note that fact that $\lambda{(\mQ)} \leq 1$, so $\lambda{(\mQ)}=1$.
\end{proof}
\subsection{Proof of Claim~\ref{claim:f}}
\label{subsec:proof_claimf}
\propertyf*
\begin{proof}
    Note that \Eqref{eq:matrix_value_func} is indeed a random value minus its expectation, so naturally \Eqref{eq:matrix_value_func} has zero mean,
    i.e., $\sum_{X\in \mathcal{S}} \sigma_X f(X) = 0$.
    Moreover, $\norm{f(X)}{2}\leq 1$ because
    \beq{
        \nonumber
        \besp{
            \norm{f(X)}{2} &\leq\frac{1}{2}\left(\sum_{r=1}^T\frac{\abs{\alpha_r}}{2}\left(\norm{\ve_{v_0}\ve^\top_{v_{r}}}{2} + \norm{\ve_{v_{r}} \ve^\top_{v_0}}{2}\right) + \sum_{r=1}^T\frac{\abs{\alpha_r}}{2} \left( \norm{\mPi}{2} \norm{\mP}{2}^r + \norm{\mP^\top}{2}^r \norm{\mPi}{2}\right)\right)\\
            &\leq  \frac{1}{2}\left(\sum_{r=1}^T \abs{\alpha_r} +  \sum_{r=1}^T \abs{\alpha_r} \right)=1.
        }}where the first step follows triangle inequaity and submultiplicativity  of 2-norm, and the third step follows by
    (1) $\norm{\ve_i \ve_j^\top}{2} = 1$;
    (2) $\norm{\mPi}{2} = \norm{\diag{\vpi}}{2} \leq  1$ for distribution $\vpi$; (3) $\norm{\mP}{2} = \norm{\mP^\top}{2}= 1$.
\end{proof}

\subsection{Proof of Corollary~\ref{col:hmm}}
\label{subsec:proof_hmm}
\hmm*
\begin{proof}
A HMM can be model by a Markov chain $\mP$ on $\mathcal{Y}\times \mathcal{X}$ such that 
$P(y_{t+1}, x_{t+1}|y_{t}, x_{t})=P(y_{t+1}|x_{t+1})P(x_{t+1}|x_{t})$. For the co-occurrence matrix of observable states, applying a similar proof like our Theorem~\ref{thm:rate} shows that one needs a trajectory of length $O(\tau(\mP) (\log{|\mathcal{Y}|} + \log{\tau(\mP)}) / \epsilon^2)$ to achieve error bound $\epsilon$ with high probability.
Moreover, the mixing time $\tau(\mP)$ is bounded by the mixing time of the Markov chain on the hidden state space~(i.e., $P(x_{t+1}|x_{t})$).  
\end{proof}

\section{Matrix Chernoff Bounds for Markov Chains}
\label{sec:chernoff_proof_all}

\subsection{Preliminaries}

\vpara{Kronecker Products}
If $\mA$ is an $M_1\times N_1$ matrix and $\mB$ is a $M_2\times N_2$ matrix,
then the Kronecker product $\mA \otimes \mB$ is the $M_2 M_1 \times N_1N_2$ block matrix such that
\beq{
    \nonumber
    \mA \otimes \mB = \begin{bmatrix}
        \mA_{1,1} \mB   & \cdots & \mA_{1,N_1}B   \\
        \vdots          & \ddots & \vdots         \\
        \mA_{M_1,1} \mB & \cdots & \mA_{M_1,N_1}B \\
    \end{bmatrix}.
}Kronecker product has the mixed-product property. If $\mA, \mB, \mC, \mD$
are matrices of such size that one can from the matrix products $\mA\mC$ and $\mB\mD$,
then  $(\mA \otimes \mB)(\mC\otimes \mD) = (\mA\mC)\otimes (\mB\mD)$.

\vpara{Vectorization}
For a matrix $\mX \in \mathbb{C}^{d\times d}$, $\vect(\mX) \in \mathbb{C}^{d^2}$ denote the
vertorization of the matrix $\mX$, s.t. $\vect(\mX)=\sum_{i\in[d]}\sum_{j\in[d]} \mX_{i,j} \ve_i \otimes \ve_j$,
which is the stack of rows of $\mX$. And there is a relationship between matrix multiplication and
Kronecker product s.t. $\vect(\mA\mX\mB)=(\mA\otimes \mB^\top)\vect(\mX)$.

\vpara{Matrices and Norms} For a matrix $\mA\in \sC^{N\times N}$,
we use $\mA^\top$ to denote matrix transpose,
use $\overline{\mA}$ to denote entry-wise matrix conjugation, use $\mA^\ast$ to denote
matrix conjugate transpose~($\mA^\ast=\overline{\mA^\top}=\overline{\mA}^\top$).
The vector 2-norm is defined to be $\norm{\vx}{2} = \sqrt{\vx^\ast \vx}$,
and the matrix 2-norm is defined to be $\norm{\mA}{2} = \max_{\vx\in \mathbb{C}^N, \vx\neq 0} \frac{\norm{\mA \vx}{2}}{\norm{\vx}{2}}$.

We then recall the definition of inner-product under $\vpi$-kernel  in \Secref{sec:preliminaries}.
The inner-product under $\vpi$-kernel for $\mathbb{C}^N$ is
$\left\langle \vx, \vy \right \rangle_\vpi = \vy^\ast \mPi^{-1}  \vx$ where $\mPi=\diag{\vpi}$,
and its induced $\vpi$-norm $\norm{\vx}{\vpi} = \sqrt{  \left\langle \vx, \vx \right \rangle_\vpi}$.
The above definition allow us to define a inner product  under $\vpi$-kernel on $\mathbb{C}^{Nd^2}$:
\begin{definition}
    Define inner product on $\sC^{Nd^2}$ under $\vpi$-kernel to be
    $\left\langle \vx, \vy \right \rangle_\vpi = \vy^\ast \left(\mPi^{-1} \otimes \mI_{d^2}\right) \vx$.
\end{definition}
\begin{remark}
    \label{rmk:inner_product_factorization}
    For $\vx, \vy \in \sC^N$ and $\vp, \vq \in \sC^{d^2}$, then inner product~(under $\vpi$-kernel)
    between $\vx \otimes \vp$ and $\vy \otimes \vq$ can be simplified as
    \beq{
        \nonumber
        \langle \vx \otimes \vp, \vy \otimes \vq \rangle_{\vpi} = (\vy \otimes \vq )^\ast \left(\mPi^{-1} \otimes \mI_{d^2}\right) (\vx \otimes \vp)
        = (\vy^\ast \mPi^{-1} \vx)\otimes (\vq^\ast \vp) = \langle \vx, \vy \rangle_{\vpi} \langle \vp, \vq\rangle.
    }
\end{remark}
\begin{remark}
    \label{rmk:norm_factorization}
    The induced $\vpi$-norm is $\norm{\vx}{\vpi} = \sqrt{  \left\langle \vx, \vx \right \rangle_\vpi}$.
    When $\vx=\vy \otimes \vw$, the $\vpi$-norm can be simplified to be:
    $            \norm{\vx}{\vpi}  =\sqrt{ \left\langle \vy \otimes \vw, \vy \otimes \vw \right \rangle_\vpi}
        = \sqrt{\langle \vy, \vy \rangle_{\vpi}\langle \vw, \vw\rangle}
        = \norm{\vy}{\vpi} \norm{\vw}{2}
    $.
\end{remark}

\vpara{Matrix Exponential} The matrix exponential of a matrix $\mA \in \sC^{d\times d}$ is defined by Taylor expansion
$\exp{(\mA)} = \sum_{j=0}^{+\infty} \frac{\mA^j}{j!}$. And we will use the fact that
$\exp(\mA) \otimes \exp(\mB) = \exp(\mA\otimes \mI + \mI \otimes \mB)$.

\vpara{Golden-Thompson Inequality} We need the following multi-matrix Golden-Thompson inequality from from \citet{garg2018matrix}.
\goldenthompson*

\subsection{Proof of Theorem~\ref{thm:chernoff}}
\label{sec:proof_chernoff}

\chernoff*

\begin{proof}
    Due to symmetry, it suffices to prove one of the statements. Let $t>0$ be a parameter to be chosen later. Then
    \beq{
        \label{eq:prob_lambda}
        \besp{
            \mathbb{P}\left[\lambda_{\max}\left( \frac{1}{k}\sum_{j=1}^k f(v_j)\right)\geq \epsilon\right] & = \mathbb{P}\left[\lambda_{\max}\left( \sum_{j=1}^k f(v_j)\right)\geq k\epsilon\right]\\
            &\leq   \mathbb{P}\left[\Tr{\left[\exp{\left( t\sum_{j=1}^k f(v_j)\right)}\right]}\geq \exp{(tk\epsilon)}\right]\\
            &\leq  \frac{\mathbb{E}_{v_1\cdots, v_k}\left[\Tr{\left[\exp{\left( t\sum_{j=1}^k f(v_j)\right)}\right]}\right]}{\exp{(tk\epsilon)}}.
        }
    }The second inequality follows Markov inequality.

    Next to bound $\mathbb{E}_{v_1\cdots, v_k}\left[\Tr{\left[\exp{\left( t\sum_{j=1}^k f(v_j)\right)}\right]}\right]$. Using Theorem~\ref{thm:golden_thompson}, we have:
    \beq{
        \nonumber
        \besp{
            \log{\left(\Tr{\left[\exp{\left( t\sum_{j=1}^k f(v_j)\right)}\right]}\right)} &\leq \frac{4}{\pi} \int_{-\frac{\pi}{2}}^{\frac{\pi}{2}}
            \log{\left( \Tr{\left[ \prod_{j=1}^k \exp{\left( \frac{e^{\iu \phi}}{2} tf(v_j)\right)} \prod_{j=k}^1 \exp{\left( \frac{e^{-\iu \phi}}{2} tf(v_j)\right)}\right]}\right)} d\mu(\phi)\\
            &\leq  \frac{4}{\pi} \log \int_{-\frac{\pi}{2}}^{\frac{\pi}{2}}
            \Tr{\left[ \prod_{j=1}^k \exp{\left( \frac{e^{\iu \phi}}{2} tf(v_j)\right)} \prod_{j=k}^1 \exp{\left( \frac{e^{-\iu \phi}}{2} tf(v_j)\right)}\right]} d\mu(\phi),
        }
    }where the second step follows by concavity of $\log$ function and the fact that $\mu(\phi)$ is a probability distribution on  $[-\frac{\pi}{2}, \frac{\pi}{2}]$.
    This implies
    \beq{    \nonumber
        \Tr{\left[\exp{\left( t\sum_{j=1}^k f(v_j)\right)}\right]} \leq \left(\int_{-\frac{\pi}{2}}^{\frac{\pi}{2}}
        \Tr{\left[ \prod_{j=1}^k \exp{\left( \frac{e^{\iu \phi}}{2} tf(v_j)\right)} \prod_{j=k}^1 \exp{\left( \frac{e^{-\iu \phi}}{2} tf(v_j)\right)}\right]} d\mu(\phi)\right)^{\frac{4}{\pi}}.
    }Note that $\norm{\vx}{p} \leq d^{1/p-1} \norm{\vx}{1}$ for $p\in (0,1)$, choosing $p=\pi/4$ we have
    \beq{    \nonumber
        \left( \Tr{\left[\exp{\left(\frac{\pi}{4} t\sum_{j=1}^k f(v_j)\right)}\right]}\right)^\frac{4}{\pi} \leq d^{\frac{4}{\pi}-1}\Tr{\left[\exp{\left( t\sum_{j=1}^k f(v_j)\right)}\right]}.
    }
    Combining the above two equations together, we have
    \beq{
        \label{eq:trace_3}
        \Tr{\left[\exp{\left(\frac{\pi}{4} t\sum_{j=1}^k f(v_j)\right)}\right]} \leq d^{1-\frac{\pi}{4}} \int_{-\frac{\pi}{2}}^{\frac{\pi}{2}}
        \Tr{\left[ \prod_{j=1}^k \exp{\left( \frac{e^{\iu \phi}}{2} tf(v_j)\right)} \prod_{j=k}^1 \exp{\left( \frac{e^{-\iu \phi}}{2} tf(v_j)\right)}\right]} d\mu(\phi).
    }
    Write  $e^{\iu \phi} = \gamma + \iu b$ with $\gamma^2+b^2 = \abs{\gamma + \iu b}^2= \abs{e^{\iu\phi}}^2  =1$:
    \garglemma*
    Assuming the above lemma, we can complete the proof of the theorem as:
    \beq{
        \label{eq:exp_trace}
        \besp{
            &\mathbb{E}_{v_1\cdots, v_k}\left[\Tr{\left[\exp{\left( \frac{\pi}{4}t\sum_{j=1}^k f(v_j)\right)}\right]}\right]\\
            \leq& d^{1-\frac{\pi}{4}} \mathbb{E}_{v_1\cdots, v_k} \left[ \int_{-\frac{\pi}{2}}^{\frac{\pi}{2}}
                \left( \Tr{\left[ \prod_{j=1}^k \exp{\left( \frac{e^{\iu \phi}}{2} tf(v_j)\right)} \prod_{j=k}^1 \exp{\left( \frac{e^{-\iu \phi}}{2} tf(v_j)\right)}\right]}\right) d\mu(\phi)\right]\\
            =& d^{1-\frac{\pi}{4}}\int_{-\frac{\pi}{2}}^{\frac{\pi}{2}}  \mathbb{E}_{v_1\cdots, v_k} \left[ \Tr{\left[ \prod_{j=1}^k \exp{\left( \frac{e^{\iu \phi}}{2} tf(v_j)\right)} \prod_{j=k}^1 \exp{\left( \frac{e^{-\iu \phi}}{2} tf(v_j)\right)}\right]} \right] d\mu(\phi)\\
            \leq&  d^{1-\frac{\pi}{4}} \int_{-\frac{\pi}{2}}^{\frac{\pi}{2}}  \norm{\vphi}{\vpi} d\exp{\left(  kt^2\abs{e^{\iu\phi}}^2\left(1+\frac{8}{1-\lambda}\right)\right)}d\mu(\phi)\\
            =&\norm{\vphi}{\vpi} d^{2-\frac{\pi}{4}}    \exp{\left(  kt^2\left(1+\frac{8}{1-\lambda}\right)\right)} \int_{-\frac{\pi}{2}}^{\frac{\pi}{2}} d\mu(\phi)\\
            =&\norm{\vphi}{\vpi} d^{2-\frac{\pi}{4}}    \exp{\left(  kt^2\left(1+\frac{8}{1-\lambda}\right)\right)}
        }}where the first step follows \Eqref{eq:trace_3}, the second step follows by swapping $\mathbb{E}$ and $\int$, the third step follows by Lemma~\ref{lemma:stoc18_lemma_4.3},
    the forth step follows by $\abs{e^{\iu\phi}}=1$, and the last step follows by $\mu$ is a probability distribution on $[-\frac{\pi}{2}, \frac{\pi}{2}]$ so $ \int_{-\frac{\pi}{2}}^{\frac{\pi}{2}} d\mu(\phi)=1$

    Finally, putting it all together:
    \beq{
        \nonumber
        \besp{
            \mathbb{P}\left[\lambda_{\max}\left( \frac{1}{k}\sum_{j=1}^k f(v_j)\right)\geq \epsilon\right] &\leq  \frac{\mathbb{E}\left[\Tr{\left[\exp{\left( t\sum_{j=1}^k f(v_j)\right)}\right]}\right]}{\exp{(tk\epsilon)}}\\
            & =  \frac{\mathbb{E}\left[\Tr{\left[\exp{\left( \frac{\pi}{4} \left(\frac{4}{\pi}t \right)\sum_{j=1}^k f(v_j)\right)}\right]}\right]}{\exp{(tk\epsilon)}}\\
            &\leq \frac{\norm{\vphi}{\vpi}d^{2-\frac{\pi}{4}}    \exp{\left(  k\left(\frac{4}{\pi}t \right)^2\left(1+\frac{8}{1-\lambda}\right)\right)}}{\exp{(tk\epsilon)}}\\
            &= \norm{\vphi}{\vpi}d^{2-\frac{\pi}{4}}  \exp{ \left(\left(\frac{4}{\pi} \right)^2 k \epsilon^2 (1-\lambda)^2 \frac{1}{36^2}\frac{9}{1-\lambda}  -k \frac{(1-\lambda)\epsilon}{36}\epsilon\right)}\\
            &\leq \norm{\vphi}{\vpi}d^{2} \exp{(-k\epsilon^2(1-\lambda)/72)} .
        }
    }where the first step follows by \Eqref{eq:prob_lambda}, the second step follows by \Eqref{eq:exp_trace}, the third step follows by choosing $t=(1-\lambda)\epsilon/36$. The only thing
    to be check is that $t=(1-\lambda)\epsilon/36$ satisfies $t\sqrt{\gamma^2 + b^2}=t \leq \frac{1-\lambda}{4\lambda}$. Recall that $\eps < 1$ and $\lambda \leq 1$, we have
    $t=\frac{(1-\lambda)\epsilon}{36}\leq \frac{1-\lambda}{4} \leq \frac{1-\lambda}{4\lambda}$.
\end{proof}

\subsection{Proof of Lemma~\ref{lemma:stoc18_lemma_4.3}}
\label{sec:proof_lemma}

\garglemma*

\begin{proof}
    Note that for $\mA, \mB  \in \mathbb{C}^{d\times d}$, $\left\langle (\mA \otimes \mB)\vect(\mI_d), \vect(\mI_d) \right\rangle = \Tr\left[\mA \mB^\top\right]$.
    By letting
    $\mA =  \prod_{j=1}^k \exp{\left(\frac{tf(v_j)(\gamma + \iu b)}{2}\right)}$ and
    $\mB =    \left(\prod_{j=k}^1 \exp{\left(\frac{tf(v_j)(\gamma - \iu b)}{2}\right)}\right)^\top=\prod_{j=1}^k \exp{\left(\frac{tf(v_j)(\gamma - \iu b)}{2}\right)}$.
    The trace term in LHS of Lemma~\ref{lemma:stoc18_lemma_4.3} becomes
    \beq{
        \label{eq:vec2}
        \besp{
            &\Tr
            \left[
                \prod_{j=1}^k \exp{\left(\frac{tf(v_j)(\gamma + \iu b)}{2}\right)}
                \prod_{j=k}^1 \exp{\left(\frac{tf(v_j)(\gamma - \iu b)}{2}\right)}
                \right] \\
            = &\left \langle
            \left(
            \prod_{j=1}^k \exp{\left(\frac{tf(v_j)(\gamma + \iu b)}{2}\right)}
            \otimes
            \prod_{j=1}^k \exp{\left(\frac{tf(v_j)(\gamma - \iu b)}{2}\right)}
            \right)
            \vect(\mI_d),  \vect(\mI_d)\right\rangle.
        }}By iteratively applying $(\mA \otimes \mB) (\mC \otimes \mD) = (\mA \mC)\otimes(\mB\mD)$, we have
    \beq{    \nonumber
        \besp{
            &\prod_{j=1}^k \exp{\left(\frac{tf(v_j)(\gamma + \iu b)}{2}\right)}
            \otimes
            \prod_{j=1}^k \exp{\left(\frac{t f(v_j)(\gamma - \iu b)}{2}\right)}\\
            =&  \prod_{j=1}^k\left(
            \exp{\left(\frac{tf(v_j)(\gamma + \iu b)}{2}\right)}
            \otimes
            \exp{\left(\frac{t f(v_j)(\gamma - \iu b)}{2}\right)}
            \right)\triangleq\prod_{j=1}^k \mM_{v_j},
        }
    }where we define
    \beq{
        \label{eq:M}
        \mM_{v_j} \triangleq \exp{\left(\frac{tf(v_j)(\gamma + ib)}{2}\right)}
        \otimes
        \exp{\left(\frac{t f(v_j)(\gamma - ib)}{2}\right)}.
    } Plug it to the trace term, we have
    \beq{\nonumber
        \Tr
        \left[
            \prod_{j=1}^k \exp{\left(\frac{tf(v_j)(\gamma + \iu b)}{2}\right)}
            \prod_{j=k}^1 \exp{\left(\frac{tf(v_j)(\gamma - \iu b)}{2}\right)}
            \right] =  \left\langle
        \left(\prod_{j=1}^k \mM_{v_j}\right)
        \vect(\mI_d), \vect(\mI_d) \right\rangle.
    }
    Next, taking expectation on \Eqref{eq:vec2} gives
    \beq{
        \label{eq:trace}
        \besp{
            &\E_{v_1,\cdots, v_k} \left[\Tr
                \left[
                    \prod_{j=1}^k \exp{\left(\frac{tf(v_j)(\gamma + ib)}{2}\right)}
                    \prod_{j=k}^1 \exp{\left(\frac{tf(v_j)(\gamma - ib)}{2}\right)}
                    \right]\right] \\
            =&\E_{v_1,\cdots, v_k}\left[
                \left\langle
                \left(\prod_{j=1}^k \mM_{v_j}\right)
                \vect(\mI_d), \vect(\mI_d) \right\rangle
                \right]\\
            =&
            \left\langle
            \E_{v_1,\cdots, v_k}\left[\prod_{j=1}^k \mM_{v_j}\right]
            \vect(\mI_d), \vect(\mI_d) \right\rangle.
        }}
    We turn to study $\E_{v_1,\cdots, v_k}\left[\prod_{j=1}^k \mM_{v_j}\right]$, which is characterized by the following lemma:

    \begin{lemma}
        \label{lemma:dp_non_stationary}
        Let $\mE\triangleq\diag{\mM_1, \mM_2, \cdots, \mM_N} \in \sC^{Nd^2\times Nd^2}$ and $\widetilde{\mP}\triangleq\mP\otimes \mI_{d^2}\in \R^{Nd^2\times Nd^2}$.
        For a random walk $(v_1, \cdots, v_k)$ such that $v_1$ is sampled from an arbitrary probability distribution $\vphi$ on $[N]$,
        $
            \E_{v_1,\cdots, v_k}\left[\prod_{j=1}^k \mM_{v_j}\right] =
            \left(\vphi \otimes \mI_{d^2} \right)^\top
        \left((\mE\widetilde{\mP})^{k-1}\mE\right)\left(\1 \otimes \mI_{d^2}\right)
        $, where $\1$ is the all-ones vector.
    \end{lemma}
    \begin{proof}(of Lemma~\ref{lemma:dp_non_stationary})
        We always treat  $\mE\widetilde{\mP} $ as a block matrix, s.t.,
        \beq{
            \nonumber
            \mE\widetilde{\mP}  =
            \begin{bmatrix}
                \mM_1 &        &       \\
                      & \ddots &       \\
                      &        & \mM_N
            \end{bmatrix}
            \begin{bmatrix}
                \mP_{1,1} \mI_{d^2} & \cdots & \mP_{1,N} \mI_{d^2} \\
                \vdots              & \ddots & \vdots              \\
                \mP_{N,1} \mI_{d^2} & \cdots & \mP_{N,N} \mI_{d^2}
            \end{bmatrix}
            =
            \begin{bmatrix}
                \mP_{1,1} \mM_1 & \cdots & \mP_{1,N} \mM_1 \\
                \vdots          & \ddots & \vdots          \\
                \mP_{N,1} \mM_N & \cdots & \mP_{N,N} \mM_N
            \end{bmatrix}.
        }I.e., the $(u, v)$-th block of $\mE\widetilde{\mP} $, denoted by $(\mE\widetilde{\mP})_{u, v}$,  is $\mP_{u, v}\mM_u$.
        \beq{
        \nonumber
        \besp{
        \E_{v_1,\cdots, v_k}\left[\prod_{j=1}^k \mM_{v_j}\right] &= \sum_{v_1, \cdots, v_k} \vphi_{v_1} \mP_{v_1, v_2}\cdots \mP_{v_{k-1}, v_k} \prod_{j=1}^k \mM_{v_j}\\
        &=  \sum_{v_1} \vphi_{v_1} \sum_{v_2}   \left(\mP_{v_1, v_2} \mM_{v_1}\right) \cdots \sum_{v_k} \left(\mP_{v_{k-1}, v_k} \mM_{v_{k-1}}\right) \mM_{v_k}\\
        & = \sum_{v_1} \vphi_{v_1} \sum_{v_2}(\mE\widetilde{\mP})_{v_1, v_2} \sum_{v_3} (\mE\widetilde{\mP})_{v_2, v_3} \cdots   \sum_{v_k} (\mE\widetilde{\mP}\mE)_{v_{k-1}, v_{k}} \\
        & = \sum_{v_1} \vphi_{v_1} \sum_{v_k} \left((\mE\widetilde{\mP})^{k-1}\mE\right)_{v_1, v_k}
        =  \left(\vphi \otimes \mI_{d^2} \right)^\top
        \left((\mE\widetilde{\mP})^{k-1}\mE\right)\left(\1 \otimes \mI_{d^2}\right)
        }
        }
    \end{proof}

    Given Lemma~\ref{lemma:dp_non_stationary}, \Eqref{eq:trace} becomes:
    \beq{
        \nonumber
        \besp{
            &\E_{v_1, \cdots, v_k} \left[\Tr
                \left[
                    \prod_{j=1}^k \exp{\left(\frac{tf(v_j)(\gamma + \iu b)}{2}\right)}
                    \prod_{j=k}^1 \exp{\left(\frac{tf(v_j)(\gamma - \iu b)}{2}\right)}
                    \right]\right] \\
            =&\left \langle  \E_{v_1,\cdots, v_k}\left[\prod_{j=1}^k M_{v_j}\right] \vect(\mI_d),
            \vect(\mI_d) \right\rangle\\
            =&\left \langle  \left(\vphi \otimes \mI_{d^2} \right)^\top
        \left((\mE\widetilde{\mP})^{k-1}\mE\right)\left(\1 \otimes \mI_{d^2}\right),
            \vect(\mI_d) \right\rangle\\
            =&\left \langle \left((\mE\widetilde{\mP})^{k-1}\mE\right) \left(\1 \otimes \mI_{d^2}\right)\vect(\mI_d),
            \left(\vphi \otimes \mI_{d^2} \right)\vect(\mI_d) \right\rangle\\
            =&\left \langle \left((\mE\widetilde{\mP})^{k-1}\mE\right) \left(\1 \otimes \vect(\mI_d)\right),
            \vpi \otimes \vect(\mI_d) \right\rangle
        }}The third equality is due to $\langle x, \mA y \rangle = \langle \mA^\ast x, y \rangle$.
    The forth equality is by setting $\mC=1$~(scalar) in $(\mA\otimes\mB)(\mC\otimes\mD)=(\mA\mC)\otimes(\mB\mD)$.
    Then
    \beq{
    \nonumber
    \besp{
    &\E_{v_1, \cdots, v_k} \left[\Tr
        \left[
            \prod_{j=1}^k \exp{\left(\frac{tf(v_j)(\gamma + \iu b)}{2}\right)}
            \prod_{j=k}^1 \exp{\left(\frac{tf(v_j)(\gamma - \iu b)}{2}\right)}
            \right]\right] \\
    =&\left \langle \left((\mE\widetilde{\mP})^{k-1}\mE\right) \left(\1 \otimes \vect(\mI_d)\right),
    \vphi \otimes \vect(\mI_d) \right\rangle\\
    =& (\vphi \otimes \vect(\mI_d))^\ast \left((\mE\widetilde{\mP})^{k-1}\mE\right) (\1 \otimes \vect(\mI_d))\\
    =&(\vphi \otimes \vect(\mI_d))^\ast \left((\mE\widetilde{\mP})^{k-1}\mE\right) \left(\left(\mP \mPi^{-1}\vpi\right) \otimes \left(\mI_{d^2}\mI_{d^2}\vect(\mI_d)\right)\right)\\
    =& (\vphi \otimes \vect(\mI_d))^\ast \left(\mE\widetilde{\mP}\right)^k \left(\mPi^{-1} \otimes \mI_{d^2}\right) (\vpi \otimes \vect(\mI_d))
    \triangleq \left\langle \vpi \otimes \vect(\mI_d), \vz_k \right\rangle_\vpi,
    }}where we define $\vz_0 = \vphi \otimes \vect(\mI_d)$  and
    $\vz_k = \left(\vz_0^\ast \left(\mE\widetilde{\mP}\right)^{k} \right)^\ast = \left(\vz_{k-1}^\ast \mE\widetilde{\mP} \right)^\ast$.
    Moreover, by Remark~\ref{rmk:norm_factorization}, we have $\norm{\vpi \otimes \vect(\mI_d)}{\vpi} = \norm{\vpi}{\vpi} \norm{\vect(\mI_d)}{2}=\sqrt{d}$ and $\norm{\vz_0}{\vpi}=\norm{\vphi \otimes \vect(\mI_d)}{\vpi} = \norm{\vphi}{\vpi} \norm{\vect(\mI_d)}{2}=\norm{\vphi}{\vpi}\sqrt{d}$
    
    \begin{definition}
        Define linear subspace 
        $\mathcal{U} = \left\{\vpi \otimes \vw, \vw \in \sC^{d^2}\right\}$.
    \end{definition}
    \begin{remark}
        $\{\vpi\otimes \ve_i, i\in [d^2]\}$ is an orthonormal basis of $\mathcal{U}$.
        This is because
        $\langle\vpi \otimes \ve_i, \vpi \otimes \ve_j \rangle_\vpi = \langle \vpi, \vpi \rangle_\vpi \langle \ve_i, \ve_j \rangle = \delta_{ij}$
        by Remark~\ref{rmk:inner_product_factorization}, where $\delta_{ij}$ is the Kronecker delta.
    \end{remark}
    \begin{remark}
        \label{rmk:projection_onto_U}
        Given $\vx=\vy \otimes \vw$. The projection of $\vx$ on to $\mathcal{U}$ is $\vx^{\parallel}=(\1^\ast \vy) (\vpi\otimes \vw)$. This is because
        \beq{
            \nonumber
            \besp{
                \vx^{\parallel} &= \sum_{i=1}^{d^2}  \langle \vy \otimes \vw, \vpi \otimes \ve_i \rangle_\vpi (\vpi \otimes \ve_i)
                =\sum_{i=1}^{d^2}  \langle \vy, \vpi \rangle_{\vpi} \langle \vw, \ve_i\rangle (\vpi \otimes \ve_i)
                = (\1^\ast \vy) (\vpi\otimes \vw).
            }
        }
    \end{remark}
    We want to bound
    \comment{
    $
        \left\langle \vz_0, \vz_k \right\rangle_{\vpi} = \left\langle \vz_0, \vz^\perp_k  + \vz^\parallel_k \right\rangle_{\vpi}
        = \left\langle \vz_0, \vz^\parallel_k \right\rangle_{\vpi} \leq \norm{\vz_0}{\vpi} \norm{\vz_k^\parallel}{\vpi} = \sqrt{d} \norm{\vz_k^\parallel}{\vpi}
    $.}
\beq{
\nonumber
\besp{
        \left\langle \vpi \otimes \vect(\mI_d), \vz_k \right\rangle_{\vpi} &= \left\langle \vpi \otimes \vect(\mI_d), \vz^\perp_k  + \vz^\parallel_k \right\rangle_{\vpi}
        = \left\langle \vpi \otimes \vect(\mI_d), \vz^\parallel_k \right\rangle_{\vpi} \\
        &\leq \norm{\vpi \otimes \vect(\mI_d)}{\vpi} \norm{\vz_k^\parallel}{\vpi} 
        = \sqrt{d} \norm{\vz_k^\parallel}{\vpi}.
}}

    As $\vz_k$ can be expressed as recursively applying operator $\mE$ and $\widetilde{\mP}$ on $\vz_0$, we turn to analyze the effects of $\mE$ and $\widetilde{\mP}$  operators.

    \begin{definition}
        \label{def:lambda}
        The spectral expansion of $\widetilde{\mP}$ is defined as
        $
            \lambda(\widetilde{\mP}) \triangleq\max_{\vx \perp \mathcal{U}, \vx\neq 0} \frac{\norm{\left(\vx^\ast \widetilde{\mP}\right)^\ast}{\vpi}}{\norm{\vx}{\vpi}}\\
        $
    \end{definition}
    \begin{lemma}
        $\lambda(\mP) =\lambda(\widetilde{\mP})$.
    \end{lemma}
    \begin{proof}

        First show  $\lambda{( \widetilde{\mP})} \geq \lambda{(\mP)}$.
        Suppose the maximizer of $ \lambda(\mP)\triangleq\max_{\vy \perp \vpi, \vy \neq 0} \frac{\norm{\left(\vy^\ast \mP\right)^\ast}{\vpi}}{\norm{\vy}{\vpi}}$
        is $\vy \in \mathbb{C}^{n}$, i.e., $\norm{\left(\vy^\ast \mP\right)^\ast}{\vpi} = \lambda(\mP)\norm{\vy}{\vpi}$.
        Construct $\vx = \vy \otimes \vo$ for arbitrary non-zero $\vo \in \mathbb{C}^{d^2}$.
        Easy to check that $\vx \perp \mathcal{U}$, because
        $\langle \vx, \vpi \otimes \vw\rangle_{\vpi} = \langle \vy , \vpi \rangle_{\vpi} \langle \vo, \vw\rangle = 0$,
        where the last equality is due to $\vy \perp \vpi$. Then we can bound $\norm{\left(\vx^\ast \widetilde{\mP}\right)^\ast}{\vpi}$ such that
        \beq{
            \nonumber
            \besp{
                \norm{\left(\vx^\ast \widetilde{\mP}\right)^\ast}{\vpi} &= \norm{\widetilde{\mP}^\ast \vx}{\vpi} = \norm{(\mP^\ast \otimes \mI_{d^2}) (\vy\otimes \vo) }{\vpi} = \norm{(\mP^\ast \vy) \otimes \vo }{\vpi}\\
                &=\norm{\left(\vy^\ast \mP\right)^\ast}{\vpi}\norm{\vo}{2}=\lambda(\mP)\norm{\vy}{\vpi}\norm{\vo}{2}=\lambda(\mP)\norm{\vx}{\vpi},
            }
        }which indicate for $\vx= \vy \otimes \vo$,
        $\frac{    \norm{  \left(\vx^\ast  \widetilde{\mP}\right)^\ast}{\vpi} }{\norm{\vx}{\vpi}} =  \lambda{(\mP)}$.
        Taking maximum over all $\vx$ gives $ \lambda{( \widetilde{\mP})} \geq  \lambda{(\mP)}$.

        Next to show  $\lambda(\mP) \geq \lambda( \widetilde{\mP})$. For $\forall \vx\in \mathbb{C}^{Nd^2}$ such that  $\vx \perp \mathcal{U}$ and $\vx \neq 0$, we can decompose it to be
        \beq{
            \nonumber
            \besp{
                \vx &= \begin{bmatrix}
                    x_1    \\
                    x_2    \\
                    \vdots \\
                    x_{Nd^2}
                \end{bmatrix} = \begin{bmatrix}
                    x_1       \\
                    x_{d^2+1} \\
                    \vdots    \\
                    x_{(N-1)d^2+1}
                \end{bmatrix} \otimes \ve_1 +
                \begin{bmatrix}
                    x_2       \\
                    x_{d^2+2} \\
                    \vdots    \\
                    x_{(N-1)d^2+2}
                \end{bmatrix} \otimes \ve_2 + \cdots +
                \begin{bmatrix}
                    x_{d^2}  \\
                    x_{2d^2} \\
                    \vdots   \\
                    x_{Nd^2}
                \end{bmatrix} \otimes \ve_{d^2}\triangleq \sum_{i=1}^{d^2} \vx_i \otimes \ve_i,
            }}where we define $\vx_i \triangleq\begin{bmatrix}x_i  &\cdots& x_{(N-1)d^2 + i} \end{bmatrix}^\top$ for $i\in[d^2]$. We can observe that $\vx_i \perp \vpi, i\in [d^2]$, because for $\forall j \in [d^2]$, we have
        \beq{
            \nonumber
            0 = \langle \vx, \vpi \otimes \ve_j\rangle_{\vpi} = \left\langle \sum_{i=1}^{d^2} \vx_i \otimes \ve_i, \vpi \otimes \ve_j\right\rangle_{\vpi}=\sum_{i=1}^{d^2}\left\langle  \vx_i \otimes \ve_i, \vpi \otimes \ve_j\right\rangle_{\vpi}
            =\sum_{i=1}^{d^2} \langle \vx_i, \vpi\rangle_{\vpi} \langle \ve_i, \ve_j \rangle= \langle \vx_j, \vpi\rangle_{\vpi},
        }which indicates $\vx_j \perp \vpi, j\in[d^2]$.
        Furthermore, we can also observe that  $\vx_i \otimes \ve_i, i\in [d^2]$  is pairwise orthogonal.
        This is because for $\forall i, j\in [d^2]$,
        $\langle\vx_i \otimes \ve_i, \vx_j \otimes \ve_j\rangle_{\vpi}
            = \langle\vx_i , \vx_j \rangle_{\vpi}\langle \ve_i, \ve_j \rangle = \delta_{ij}
        $, which suggests us to use Pythagorean theorem such that
        $\norm{\vx}{\vpi}^2 = \sum_{i=1}^{d^2} \norm{\vx_i \otimes \ve_i}{\vpi}^2 =  \sum_{i=1}^{d^2} \norm{\vx_i}{\vpi} \norm{\ve_i}{2}^2$.

        We can use similar way to decompose and analyze $\left(\vx^\ast  \widetilde{\mP}\right)^\ast$:
        \beq{\nonumber
            \left(\vx^\ast  \widetilde{\mP}\right)^\ast =  \widetilde{\mP}^\ast \vx =\sum_{i=1}^{d^2}   (\mP^\ast \otimes \mI_{d^2}) (\vx_i \otimes \ve_i)=\sum_{i=1}^{d^2} (\mP^\ast \vx_i) \otimes \ve_i.
        }where we can observe that  $(\mP^\ast \vx_i) \otimes \ve_i, i\in [d^2]$  is pairwise orthogonal. This is because for $\forall i, j\in [d^2]$, we have $
            \langle  (\mP^\ast \vx_i)  \otimes \ve_i,  (\mP^\ast \vx_j)  \otimes \ve_j\rangle_{\vpi} = \langle \mP^\ast \vx_i, \mP^\ast \vx_j\rangle_{\vpi} \langle \ve_i, \ve_j \rangle = \delta_{ij}
        $. Again, applying  Pythagorean theorem gives:
        \beq{
            \nonumber
            \besp{
                \norm{  \left(\vx^\ast  \widetilde{\mP}\right)^\ast}{\vpi}^2 &= \sum_{i=1}^{d^2} \norm{(\mP^\ast \vx_i) \otimes \ve_i}{\vpi}^2= \sum_{i=1}^{d^2} \norm{\left(\vx_i^\ast\mP\right)^\ast  }{\vpi}^2 \norm{\ve_i}{2}^2\\
                &\leq \sum_{i=1}^{d^2} \lambda{(\mP)}^2\norm{\vx_i  }{\vpi}^2 \norm{\ve_i}{2}^2= \lambda{(\mP)}^2 \left( \sum_{i=1}^{d^2} \norm{\vx_i  }{\vpi}^2 \norm{\ve_i}{2}^2\right)= \lambda{(\mP)}^2\norm{\vx}{\vpi}^2,
            }}which indicate that for $\forall \vx$ such that $\vx\perp \mathcal{U}$ and $\vx \neq 0$, we have
        $
            \frac{    \norm{  \left(\vx^\ast  \widetilde{\mP}\right)^\ast}{\vpi} }{\norm{\vx}{\vpi}} \leq  \lambda{(\mP)}
        $, or equivalently  $ \lambda{( \widetilde{\mP})} \leq  \lambda{(\mP)}$.

        Overall, we have shown both $\lambda{( \widetilde{\mP})} \geq  \lambda{(\mP)}$ and $\lambda{( \widetilde{\mP})} \leq  \lambda{(\mP)}$.
        We conclude $ \lambda{( \widetilde{\mP})} =  \lambda{(\mP)}$.
    \end{proof}

    \begin{lemma}{\textbf{(The effect of $\widetilde{\mP}$ operator)}}
        \label{lemma:effect_of_p}
        This lemma is a generalization of lemma~3.3 in \cite{chung2012chernoff}.
        \begin{enumerate}
            \item $\forall \vy \in \mathcal{U}$, then  $\left(\vy^\ast \widetilde{\mP}\right)^\ast = \vy $.
            \item $\forall \vy \perp \mathcal{U}$, then  $\left(\vy^\ast\widetilde{\mP}\right)^\ast \perp \mathcal{U}$, and $\norm{\left(\vy^\ast \widetilde{\mP}\right)^\ast}{\vpi} \leq \lambda \norm{\vy}{\vpi}$.
        \end{enumerate}
    \end{lemma}
    \begin{proof}
        First prove the Part 1 of lemma~\ref{lemma:effect_of_p}.
        $\forall \vy = \vpi \otimes \vw \in \mathcal{U}$:
        \beq{\nonumber
            \vy^\ast \widetilde{\mP} = \left(\vpi^\ast \otimes \vw^\ast\right)(\mP\otimes \mI_{d^2})
            = (\vpi^\ast \mP) \otimes \left(\vw^\ast \mI_{d^2} \right)
            = \vpi^\ast \otimes  \vw^\ast
            = \vy^\ast,
        }where third equality is becase $\vpi$ is the stationary distribution.
        Next to prove Part 2 of lemma~\ref{lemma:effect_of_p}. Given $\vy \perp \mathcal{U}$, want to show
        $(\vy^\ast\widetilde{\mP})^\ast \perp \vpi\otimes \vw $, for every $\vw \in \sC^{d^2}$.
        It is true because
        \beq{
            \nonumber
            \besp{
                \left\langle  \vpi\otimes \vw, (\vy^\ast\widetilde{\mP})^\ast  \right\rangle_\vpi
                =& \vy^\ast \widetilde{\mP} \left(\mPi^{-1} \otimes \mI_{d^2}\right) ( \vpi\otimes \vw)
                = \vy^\ast \left((\mP\mPi^{-1}\vpi) \otimes \vw\right)
                = \vy^\ast \left((\mPi^{-1}\vpi) \otimes \vw\right)\\
                =& \vy^\ast\left(\mPi^{-1} \otimes \mI_{d^2}\right) (\vpi \otimes \vw)
                = \langle  \vpi \otimes \vw, \vy \rangle_\vpi = 0.
            }
        }
        The third equality is due to $\mP\mPi^{-1}\vpi=\mP\1=\1=\mPi^{-1}\vpi$.
        Moreover, $\norm{\left(\vy^\ast \widetilde{\mP}\right)^\ast}{\vpi} \leq \lambda \norm{\vy}{\vpi}$ is simply a re-statement of definition~\ref{def:lambda}.
    \end{proof}
    \begin{remark} 
    \label{rmk:effect_of_p}
    Lemma~\ref{lemma:effect_of_p} implies that $\forall \vy \in \sC^{nd^2}$
    \begin{enumerate}
    \item $
         \left(\left(\vy^\ast \widetilde{\mP}\right)^{\ast}\right)^\parallel =  
           \left(\left(\vy^{\parallel\ast} \widetilde{\mP}\right)^{\ast}\right)^\parallel
           +            \left(\left(\vy^{\perp\ast} \widetilde{\mP}\right)^{\ast}\right)^\parallel = \vy^\parallel + \bm{0} = \vy^\parallel$
    \item $
    \left(\left(\vy^\ast
    \widetilde{\mP}\right)^{\ast}\right)^\perp =  
           \left(\left(\vy^{\parallel\ast} \widetilde{\mP}\right)^{\ast}\right)^\perp
           +            \left(\left(\vy^{\perp\ast} \widetilde{\mP}\right)^{\ast}\right)^\perp  = \bm{0} + \left(\vy^{\perp\ast} \widetilde{\mP}\right)^{\ast} = \left(\vy^{\perp\ast} \widetilde{\mP}\right)^{\ast}$.
    \end{enumerate}
    \end{remark}


    \begin{lemma}{\textbf{(The effect of $\mE$ operator)}}
        \label{lemma:effect_of_E_new}
        Given three parameters $\lambda \in [0, 1], \ell \geq 0$ and $t > 0$. Let $\mP$ be a regular Markov chain on state space $[N]$,
        with stationary distribution $\vpi$ and
        spectral expansion $\lambda$. Suppose each state $i\in[N]$ is assigned a matrix $\mH_i \in \sC^{d^2\times d^2}$ s.t. $\norm{\mH_i}{2} \leq \ell$ and
        $\sum_{i\in[N]} \pi_i \mH_i =0$. Let $\widetilde{\mP} = \mP \otimes \mI_{d^2}$ and $\mE$ denotes the $Nd^2 \times  Nd^2$ block matrix where the $i$-th
        diagonal block is the matrix $\exp{(t\mH_i)}$, i.e., $\mE=\diag{\exp{(t\mH_1)}, \cdots, \exp{(t\mH_N)}}$.
        Then for any $\forall \vz \in \sC^{Nd^2}$, we have:
        \begin{enumerate}
            \item $\norm{\left( \left(\vz^{\parallel\ast} \mE\widetilde{\mP}\right)^\ast \right)^{\parallel}}{\vpi}\leq \alpha_1 \norm{\vz^\parallel}{\vpi}$, where $\alpha_1=\exp{(t\ell)} - t\ell$.
            \item $\norm{\left( \left(\vz^{\parallel\ast} \mE\widetilde{\mP}\right)^\ast \right)^{\perp}}{\vpi}\leq \alpha_2 \norm{\vz^\parallel}{\vpi}$, where $\alpha_2= \lambda(\exp{(t\ell)} - 1)$.
            \item $\norm{ \left(\left(\vz^{\perp\ast}  \mE\widetilde{\mP}\right)^\ast\right)^\parallel }{\vpi}  \leq \alpha_3 \norm{\vz^{\perp} }{\vpi}$, where $\alpha_3= \exp{(t\ell)} - 1$.
            \item $\norm{ \left(\left(\vz^{\perp\ast}  \mE\widetilde{\mP}\right)^\ast\right)^\perp }{\vpi} \leq \alpha_4 \norm{\vz^\perp}{\vpi}$, where $\alpha_4=\lambda\exp{(t \ell)} $.
        \end{enumerate}
    \end{lemma}
    \begin{proof}(of Lemma~\ref{lemma:effect_of_E_new})
        We first show that, for $\vz=\vy\otimes \vw$,
        \beq{\nonumber
            \besp{
                \left(\vz^\ast \mE\right)^\ast = \mE^\ast \vz &=
                \begin{bmatrix}
                    \exp(t\mH^\ast_1) &        &                   \\
                                      & \ddots &                   \\
                                      &        & \exp(t\mH^\ast_N)
                \end{bmatrix}
                \begin{bmatrix}
                    y_1 \vw \\
                    \vdots  \\
                    y_N \vw
                \end{bmatrix}
                = \begin{bmatrix}
                    y_1    \exp(t\mH^\ast_1)  \vw \\
                    \vdots                        \\
                    y_N     \exp(t\mH^\ast_N) \vw
                \end{bmatrix} \\
                &=   \begin{bmatrix}
                    y_1    \exp(t\mH^\ast_1)  \vw \\
                    \vdots                        \\
                    0
                \end{bmatrix}  + \cdots +
                \begin{bmatrix}
                    0      \\
                    \vdots \\
                    y_N     \exp(t\mH^\ast_N) \vw
                \end{bmatrix}
                = \sum_{i=1}^N y_i \left(\ve_i \otimes (\exp(t\mH^\ast_i)\vw)\right).
            }
        }Due to the linearity of projection,
        \beq{
            \label{eq:E_projection}
            \besp{
                \left( \left(\vz^\ast \mE\right)^\ast \right)^{\parallel} &=\sum_{i=1}^N y_i \left(\ve_i \otimes (\exp(t\mH^\ast_i)\vw)\right)^{\parallel}
                =  \sum_{i=1}^N y_i (\1^\ast\ve_i) \left(\vpi \otimes  (\exp(t\mH^\ast_i)\vw)\right)
                = \vpi \otimes  \left(\sum_{i=1}^N y_i \exp(t\mH^\ast_i)\vw \right),
            }
        }where the second inequality follows by Remark~\ref{rmk:projection_onto_U}.

        \vpara{Proof of Lemma~\ref{lemma:effect_of_E_new}, Part 1}
        Firstly We can bound $\norm{\sum_{i=1}^N \pi_i \exp(t\mH^\ast_i) }{2}$ by
        \beq{
            \nonumber
            \besp{
                \norm{\sum_{i=1}^N \pi_i \exp(t\mH^\ast_i) }{2} &= \norm{\sum_{i=1}^N \pi_i \exp(t\mH_i) }{2}
                =\norm{\sum_{i=1}^N \pi_i \sum_{k=0}^{+\infty} \frac{t^j \mH_i^j}{j!} }{2}
                =\norm{\mI + \sum_{i=1}^N \pi_i \sum_{j=2}^{+\infty} \frac{t^j \mH_i^j}{j!}}{2}\\
                &\leq 1 + \sum_{i=1}^N \pi_i \sum_{j=2}^{+\infty} \frac{t^j\norm{\mH_i}{2}^j}{j!}
                \leq 1 + \sum_{i=1}^N \pi_i \sum_{j=2}^{+\infty} \frac{(t\ell)^j}{j!}
                = \exp{(t\ell)} - t\ell,
            }
        }where the first step follows by $\norm{\mA}{2} = \norm{\mA^\ast}{2}$, the second step follows by matrix exponential,
        the third step follows by $\sum_{i\in[N]} \pi_i \mH_i =0$, and the forth step follows by triangle inequality.
        Given the above bound, for any $\vz^\parallel$ which can be written as $\vz^\parallel=\vpi \otimes \vw$ for some $\vw \in \sC^{d^2}$, we have
        \beq{
            \nonumber
            \besp{
                \norm{\left( \left(\vz^{\parallel\ast} \mE\widetilde{\mP}\right)^\ast \right)^{\parallel}}{\vpi}
                &= \norm{\left( \left(\vz^{\parallel\ast} \mE\right)^\ast \right)^{\parallel}}{\vpi}
                =  \norm{\vpi \otimes  \left(\sum_{i=1}^N \pi_i \exp(t\mH^\ast_i)\vw \right)}{\vpi}
                = \norm{\vpi}{\vpi} \norm{\sum_{i=1}^N \pi_i \exp(t\mH^\ast_i)\vw }{2}\\
                &\leq \norm{\vpi}{\vpi} \norm{\sum_{i=1}^N \pi_i \exp(t\mH^\ast_i) }{2} \norm{\vw}{2}
                =\norm{\sum_{i=1}^N \pi_i \exp(t\mH^\ast_i) }{2} \norm{\vz^\parallel}{\vpi}\\
                &\leq \left(\exp{(t\ell)} - t\ell\right)\norm{\vz^\parallel}{\vpi},
            }
        }where step one follows by Part 1 of Remark~\ref{rmk:effect_of_p} and  step two follows by \Eqref{eq:E_projection}.

        \vpara{Proof of Lemma~\ref{lemma:effect_of_E_new}, Part 2}
        For $\forall \vz \in \mathbb{C}^{Nd^2}$, we can write it as block matrix such that:
        \beq{
            \nonumber
            \vz =   \begin{bmatrix}
                \vz_1  \\
                \vdots \\
                \vz_N
            \end{bmatrix} =
            \begin{bmatrix}
                \vz_1  \\
                \vdots \\
                0
            \end{bmatrix} + \cdots +
            \begin{bmatrix}
                0      \\
                \vdots \\
                \vz_N
            \end{bmatrix} =\sum_{i=1}^N \ve_i \otimes \vz_i,
        }where each $\vz_i\in\sC^{d^2}$. Please note that above decomposition is pairwise orthogonal. Applying Pythagorean theorem  gives
        $\norm{\vz}{\vpi}^2= \sum_{i=1}^N \norm{\ve_i \otimes \vz_i}{\vpi}^2 = \sum_{i=1}^N \norm{\ve_i}{\vpi}^2 \norm{\vz_i}{2}^2$.
        Similarly, we can decompose $(\mE^\ast - \mI_{Nd^2} ) \vz$ such that
        \beq{
            \label{eq:E_I}
            \besp{
                (\mE^\ast - \mI_{Nd^2} ) \vz &=             \begin{bmatrix}
                    \exp(t\mH^\ast_1) - \mI_{d^2} &        &                               \\
                                                  & \ddots &                               \\
                                                  &        & \exp(t\mH^\ast_N) - \mI_{d^2}
                \end{bmatrix}
                \begin{bmatrix}
                    \vz_1  \\
                    \vdots \\
                    \vz_N
                \end{bmatrix}
                =             \begin{bmatrix}
                    (\exp(t\mH^\ast_1) - \mI_{d^2}) \vz_1 \\
                    \vdots                                \\
                    (\exp(t\mH^\ast_N) - \mI_{d^2})\vz_N
                \end{bmatrix}\\
                &=\begin{bmatrix}
                    (\exp(t\mH^\ast_1) - \mI_{d^2}) \vz_1 \\
                    \vdots                                \\
                    0
                \end{bmatrix} + \cdots +
                \begin{bmatrix}
                    0      \\
                    \vdots \\
                    (\exp(t\mH^\ast_N) - \mI_{d^2})\vz_N
                \end{bmatrix}\\
                &= \sum_{i=1}^N \ve_i \otimes \left( (\exp(t\mH^\ast_i) - \mI_{d^2})\vz_i \right).
            }
        }Note that above decomposition is pairwise orthogonal, too. Applying Pythagorean theorem gives
        \beq{
            \nonumber
            \besp{
                \norm{ (\mE^\ast - \mI_{Nd^2} ) \vz}{\vpi}^2 &= \sum_{i=1}^N \norm{\ve_i \otimes \left( (\exp(t\mH^\ast_i) - \mI_{d^2})\vz_i \right)}{\vpi}^2
                = \sum_{i=1}^N \norm{\ve_i}{\vpi}^2 \norm{ (\exp(t\mH^\ast_i) - \mI_{d^2})\vz_i }{2}^2\\
                &\leq \sum_{i=1}^N \norm{\ve_i}{\vpi}^2 \norm{\exp(t\mH^\ast_i) - \mI_{d^2}}{2}^2 \norm{\vz_i }{2}^2
                \leq \max_{i\in[N]} \norm{\exp(t\mH^\ast_i) - \mI_{d^2}}{2}^2 \sum_{i=1}^N \norm{\ve_i}{\vpi}^2 \norm{\vz_i }{2}^2\\
                &=\max_{i\in[N]} \norm{\exp(t\mH^\ast_i) - \mI_{d^2}}{2}^2 \norm{\vz}{\vpi}^2=\max_{i\in[N]} \norm{\exp(t\mH_i) - \mI_{d^2}}{2}^2 \norm{\vz}{\vpi}^2,
            }
        }which indicates
        \beq{
            \nonumber
            \besp{
                \norm{ (\mE^\ast - \mI_{Nd^2} ) \vz}{\vpi}
                &=\max_{i\in[N]} \norm{\exp(t\mH_i) - \mI_{d^2}}{2} \norm{\vz}{\vpi}
                =\max_{i\in[N]} \norm{\sum_{j=1}^{+\infty} \frac{t^j \mH_i^j}{j!}}{2}\norm{\vz}{\vpi}\\
                &\leq \left(\sum_{j=1}^{+\infty} \frac{t^j\ell^j}{j!}\right)\norm{\vz}{\vpi} = (\exp{(t\ell)} - 1)\norm{\vz}{\vpi}.
            }}Now we can formally prove Part 2 of Lemma~\ref{lemma:effect_of_E_new} by:
        \beq{
            \nonumber
            \besp{
                \norm{\left( \left(\vz^{\parallel\ast} \mE\widetilde{\mP}\right)^\ast \right)^{\perp}}{\vpi} &
                =  \norm{\left(\left(\mE^\ast \vz^\parallel\right)^{\perp\ast} \widetilde{\mP}\right)^\ast}{\vpi} \leq \lambda\norm{\left(\mE^\ast \vz^\parallel\right)^{\perp}}{\vpi}
                =\lambda\norm{\left(\mE^\ast \vz^\parallel - \vz^\parallel + \vz^\parallel \right)^{\perp}}{\vpi}\\
                &  =\lambda\norm{\left( \left(\mE^\ast- \mI_{Nd^2}\right) \vz^\parallel \right)^{\perp}}{\vpi}
                \leq \lambda \norm{\left(\mE^\ast- \mI_{Nd^2}\right) \vz^\parallel }{\vpi}
                \leq \lambda(\exp{(t\ell)} - 1) \norm{\vz^\parallel}{\vpi}.
            }
        }The first step follows by Part 2 of Remark~\ref{rmk:effect_of_p}, the second step follows by Part 1 on Lemma~\ref{lemma:effect_of_p} and the forth step is due to $\left(\vz^\parallel\right)^\perp = \bm{0}$.
        
        \vpara{Proof of Lemma~\ref{lemma:effect_of_E_new}, Part 3} Note that
        \beq{
            \nonumber
            \besp{
                \norm{ \left(\left(\vz^{\perp\ast} \mE\widetilde{\mP} \right)^\ast\right)^\parallel }{\vpi} &
                = \norm{ \left(\mE^\ast\vz^\perp\right)^\parallel}{\vpi} = \norm{ \left(\mE^\ast\vz^\perp - \vz^\perp + \vz^\perp\right)^\parallel}{\vpi} 
                =\norm{ \left((\mE^\ast -  \mI_{Nd^2})\vz^\perp  \right)^\parallel}{\vpi}\\
                &\leq \norm{ (\mE^\ast -  \mI_{Nd^2})\vz^\perp  }{\vpi} \leq (\exp{(t\ell)} - 1)\norm{\vz^\perp}{\vpi},
            }
        }where the first step follows by Part 1 of Remark~\ref{rmk:effect_of_p}, the third step follows by $\left(\vz^\perp\right)^\parallel = \bm{0}$,
        and the last step follows by Part 2 of Lemma~\ref{lemma:effect_of_p}.
        
        \vpara{Proof of Lemma~\ref{lemma:effect_of_E_new}, Part 4} 
                Simiar to \Eqref{eq:E_I},  for $\forall \vz \in \mathbb{C}^{Nd^2}$, we can decompose $\mE^\ast \vz$ as
        $\mE^\ast \vz = \sum_{i=1}^N \ve_i \otimes (\exp(t\mH^\ast_i)\vz_i )$. This decomposition is pairwise orthogonal, too.
        Applying  Pythagorean theorem gives
        \beq{
            \nonumber
            \besp{
                \norm{ \mE^\ast \vz}{\vpi}^2 &= \sum_{i=1}^N \norm{\ve_i \otimes \left( \exp(t\mH^\ast_i)\vz_i \right)}{\vpi}^2
                = \sum_{i=1}^N \norm{\ve_i}{\vpi}^2 \norm{ \exp(t\mH^\ast_i) \vz_i }{2}^2
                \leq \sum_{i=1}^N \norm{\ve_i}{\vpi}^2 \norm{\exp(t\mH^\ast_i)}{2}^2 \norm{\vz_i }{2}^2\\
                &\leq \max_{i\in[N]} \norm{\exp(t\mH^\ast_i) }{2}^2 \sum_{i=1}^N \norm{\ve_i}{\vpi}^2 \norm{\vz_i }{2}^2
                \leq  \max_{i\in[N]} \exp{\left(\norm{t\mH^\ast_i }{2}\right)}^2 \norm{\vz}{\vpi}^2
                \leq \exp{(t\ell)}^2  \norm{\vz}{\vpi}^2
            }
        }which indicates $ \norm{ \mE^\ast \vz}{\vpi} \leq \exp{(t\ell)} \norm{\vz}{\vpi}$.
        Now we can prove Part 4 of Lemma~\ref{lemma:effect_of_E_new}:
        Note that
        \beq{
            \nonumber
            \besp{
                \norm{\left( \left(\vz^{\perp\ast} \mE\widetilde{\mP}\right)^\ast \right)^{\perp}}{\vpi} &
                =  \norm{\left(\left(\mE^\ast \vz^\perp\right)^{\perp\ast} \widetilde{\mP}\right)^\ast}{\vpi} \leq \lambda\norm{\left(\mE^\ast \vz^\perp\right)^{\perp}}{\vpi}
                \leq \lambda\norm{\mE^\ast \vz^\perp}{\vpi}
                \leq \lambda \exp{(t\ell)} \norm{ \vz^\perp}{\vpi}.
            }
        }
    \end{proof}

    \vpara{Recursive Analysis} We now use Lemma~\ref{lemma:effect_of_E_new} to analyze the evolution of $\vz_i^\parallel$ and $\vz_i^\perp$.
    Let  $\mH_v \triangleq \frac{f(v)(\gamma + \iu  b)}{2} \otimes \mI_{d^2} +  \mI_{d^2} \otimes \frac{f(v)(\gamma - \iu  b)}{2}$  in Lemma~\ref{lemma:effect_of_E_new}.
    We can see verify the following three facts: (1) $\exp(t\mH_v) = \mM_v$; (2) $\norm{\mH_v}{2}$ is bounded (3) $\sum_{v\in[N]}\pi_v \mH_v=0$.

    Firstly, easy to see that
    \beq{
        \nonumber
        \besp{
            \exp{(t\mH_v)} &= \exp{\left(\frac{tf(v)(\gamma + \iu  b)}{2} \otimes \mI_{d^2} +  \mI_{d^2} \otimes \frac{tf(v)(\gamma - \iu  b)}{2}\right)}\\
            &= \exp{\left(\frac{tf(v)(\gamma + \iu  b)}{2}\right)} \otimes \exp{\left( \frac{tf(v)(\gamma - \iu  b)}{2}\right)}=\mM_v,
        }}where the first step follows by definition of $\mH_i$ and the second step follows by the fact that
    $\exp(\mA\otimes \mI_d + \mI_d \otimes \mB) = \exp(\mA) \otimes \exp(\mB)$, and the last step follows by \Eqref{eq:M}.

    Secondly, we can bound $\norm{\mH_v}{2}$ by:
    \beq{
        \nonumber
        \besp{
            \norm{\mH_v}{2} &\leq  \norm{\frac{f(v)(\gamma + \iu  b)}{2} \otimes \mI_{d^2}}{2} + \norm{\mI_{d^2} \otimes \frac{f(v)(\gamma - \iu  b)}{2}}{2}\\
            &=\norm{\frac{f(v)(\gamma + \iu  b)}{2} }{2} \norm{\mI_{d^2}}{2} + \norm{\mI_{d^2}}{2}\norm{\frac{f(v)(\gamma - \iu  b)}{2} }{2}\leq \sqrt{\gamma^2 + b^2},
        }}where the first step follows by triangle inequality, the second step follows by the fact that $\norm{\mA\otimes \mB}{2} = \norm{\mA}{2} \norm{\mB}{2}$, the third step
    follows by $\norm{\mI_d}{2}=1$ and $\norm{f(v)}{2} \leq 1$. We set $\ell =\sqrt{\gamma^2 + b^2}$ to satisfy the assumption in Lemma~\ref{lemma:effect_of_E_new} that $\norm{\mH_v}{2} \leq \ell$.
    According to the conditions in Lemma~\ref{lemma:stoc18_lemma_4.3}, we know that $t\ell \leq 1$ and $t\ell \leq \frac{1-\lambda}{4\lambda}$.

    Finally, we show that $\sum_{v\in[N]}\pi_v \mH_v=0$, because
    \beq{
        \nonumber
        \besp{
            \sum_{v\in [N]}\pi_v \mH_v &= \sum_{v\in [N]} \left(\frac{f(v)(\gamma + \iu  b)}{2} \otimes \mI_{d^2} +  \mI_{d^2} \otimes \frac{f(v)(\gamma - \iu  b)}{2}\right)\\
            &= \frac{\gamma + \iu b}{2} \left(\sum_{v\in [N]} \pi_v f(v)\right) \otimes \mI_d + \frac{\gamma - \iu b}{2} \mI_d \otimes \left(\sum_{v\in[N]} \pi_v f(v)\right) = 0,
        }}where the last step follows by $\sum_{v\in[N]} \pi_v f(v)= 0$.

    \begin{claim}
        \label{claim:perp}
        $\norm{\vz_i^\perp}{\vpi} \leq \frac{\alpha_2}{1-\alpha_4} \max_{0 \leq j < i}\norm{\vz_{j}^\parallel}{\vpi}$.
    \end{claim}
    \begin{proof} Using Part 2 and Part 4 of Lemma~\ref{lemma:effect_of_E_new}, we have
    \comment{ 
        \beq{
            \nonumber
            \besp{
                \norm{\vz_i^\perp}{\vpi} &= \norm{ \left(\left(\vz_{i-1}^\ast \widetilde{\mP} \mE\right)^\ast\right)^\perp }{\vpi} \\
                &\leq \norm{ \left(\left(\vz_{i-1}^{\parallel \ast} \widetilde{\mP} \mE\right)^\ast\right)^\perp }{\vpi} +  \norm{ \left(\left(\vz_{i-1}^{\perp \ast} \widetilde{\mP} \mE\right)^\ast\right)^\perp }{\vpi} \\
                &\leq \alpha_2 \norm{\vz_{i-1}^\parallel}{\vpi} + \alpha_4 \norm{\vz_{i-1}^\perp}{\vpi}\\
                &\leq (\alpha_2 + \alpha_2 \alpha_4 + \alpha_2 \alpha_4^2 + \cdots ) \max_{0 \leq j < i}\norm{\vz_{j}^\parallel}{\vpi} \\
                & \leq \frac{\alpha_2}{1-\alpha_4} \max_{0 \leq j < i}\norm{\vz_{j}^\parallel}{\vpi}
            }}
    }
    \beq{
            \nonumber
            \besp{
                \norm{\vz_i^\perp}{\vpi} &= \norm{ \left(\left(\vz_{i-1}^\ast \mE\widetilde{\mP} \right)^\ast\right)^\perp }{\vpi} \\
                &\leq \norm{ \left(\left(\vz_{i-1}^{\parallel \ast} \mE\widetilde{\mP} \right)^\ast\right)^\perp }{\vpi} +  \norm{ \left(\left(\vz_{i-1}^{\perp \ast} \mE\widetilde{\mP} \right)^\ast\right)^\perp }{\vpi} \\
                &\leq \alpha_2 \norm{\vz_{i-1}^\parallel}{\vpi} + \alpha_4 \norm{\vz_{i-1}^\perp}{\vpi}\\
                &\leq (\alpha_2 + \alpha_2 \alpha_4 + \alpha_2 \alpha_4^2 + \cdots ) \max_{0 \leq j < i}\norm{\vz_{j}^\parallel}{\vpi} \\
                & \leq \frac{\alpha_2}{1-\alpha_4} \max_{0 \leq j < i}\norm{\vz_{j}^\parallel}{\vpi}
            }}
    \end{proof}
    \begin{claim}
        \label{claim:parallel}
        $\norm{\vz_i^\parallel}{\vpi} \leq \left(\alpha_1 + \frac{\alpha_2\alpha_3}{1-\alpha_4}\right) \max_{0 \leq j < i}\norm{\vz_{j}^\parallel}{\vpi}$.
    \end{claim}
    \begin{proof} Using Part 1 and Part 3 of Lemma~\ref{lemma:effect_of_E_new} as well as Claim~\ref{claim:perp}, we have
    \comment{ 
        \beq{
            \nonumber
            \besp{
                \norm{\vz_i^\parallel}{\vpi} &= \norm{ \left(\left(\vz_{i-1}^\ast \widetilde{\mP} \mE\right)^\ast\right)^\parallel }{\vpi}\\
                &\leq \norm{ \left(\left(\vz_{i-1}^{\parallel \ast} \widetilde{\mP} \mE\right)^\ast\right)^\parallel }{\vpi} +  \norm{ \left(\left(\vz_{i-1}^{\perp \ast} \widetilde{\mP} \mE\right)^\ast\right)^\parallel }{\vpi} \\
                &\leq \alpha_1\norm{\vz_{i-1}^\parallel}{\vpi} +\alpha_3 \norm{\vz_{i-1}^\perp}{\vpi} \\
                &\leq \alpha_1\norm{\vz_{i-1}^\parallel}{\vpi} +\alpha_3\frac{\alpha_2}{1-\alpha_4} \max_{0 \leq j < i-1}\norm{\vz_{j}^\parallel}{\vpi}\\
                &\leq \left(\alpha_1 + \frac{\alpha_2\alpha_3}{1-\alpha_4}\right) \max_{0 \leq j < i}\norm{\vz_{j}^\parallel}{\vpi}.
            }}
    }
    \beq{
            \nonumber
            \besp{
                \norm{\vz_i^\parallel}{\vpi} &= \norm{ \left(\left(\vz_{i-1}^\ast\mE\widetilde{\mP} \right)^\ast\right)^\parallel }{\vpi}\\
                &\leq \norm{ \left(\left(\vz_{i-1}^{\parallel \ast} \mE\widetilde{\mP} \right)^\ast\right)^\parallel }{\vpi} +  \norm{ \left(\left(\vz_{i-1}^{\perp \ast} \mE\widetilde{\mP} \right)^\ast\right)^\parallel }{\vpi} \\
                &\leq \alpha_1\norm{\vz_{i-1}^\parallel}{\vpi} +\alpha_3 \norm{\vz_{i-1}^\perp}{\vpi} \\
                &\leq \alpha_1\norm{\vz_{i-1}^\parallel}{\vpi} +\alpha_3\frac{\alpha_2}{1-\alpha_4} \max_{0 \leq j < i-1}\norm{\vz_{j}^\parallel}{\vpi}\\
                &\leq \left(\alpha_1 + \frac{\alpha_2\alpha_3}{1-\alpha_4}\right) \max_{0 \leq j < i}\norm{\vz_{j}^\parallel}{\vpi}.
            }}
    \end{proof}

    Combining Claim~\ref{claim:perp} and Claim~\ref{claim:parallel} gives
    \beq{
        \nonumber
        \besp{
            \norm{\vz_k^\parallel}{\vpi} &\leq \left(\alpha_1 + \frac{\alpha_2\alpha_3}{1-\alpha_4}\right) \max_{0 \leq j < k}\norm{\vz_{j}^\parallel}{\vpi}\\
            \text{(because $\alpha_1 + \alpha_2\alpha_3/(1-\alpha_4) \geq \alpha_1 \geq 1$ )}&\leq  \left(\alpha_1 + \frac{\alpha_2\alpha_3}{1-\alpha_4}\right)^k \norm{\vz_0^\parallel}{\vpi}\\
            &=\norm{\vphi}{\vpi}\sqrt{d}\left(\alpha_1 + \frac{\alpha_2\alpha_3}{1-\alpha_4}\right)^k,
        }}which implies
    \beq{\nonumber
    \left\langle \vpi \otimes \vect(\mI_d), \vz_k \right\rangle_{\vpi}
         \leq\norm{\vphi}{\vpi} d\left(\alpha_1 + \frac{\alpha_2\alpha_3}{1-\alpha_4}\right)^k.
        }
    Finally, we bound $\left(\alpha_1 + \frac{\alpha_2\alpha_3}{1-\alpha_4}\right)^k$. The same as \cite{garg2018matrix}, we can bound
    $\alpha_1, \alpha_2\alpha_3, \alpha_4$ by:
    \beq{
        \nonumber
        \alpha_1 = \exp{(t\ell)} - t\ell \leq 1+ t^2\ell^2 = 1+t^2(\gamma^2 + b^2),
    }and
    \beq{
        \nonumber
        \alpha_2\alpha_3 = \lambda (\exp{(t\ell)}-1)^2 \leq \lambda (2t\ell)^2 =4\lambda t^2(\gamma^2+b^2)
    }where the second step is because $\exp{(x)} \leq 1+2x, \forall x \in [0, 1]$ and $t\ell < 1$,
    \beq{
        \nonumber
        \alpha_4 =\lambda \exp{(t\ell)}\leq \lambda (1+2t\ell) \leq \frac{1}{2}+\frac{1}{2}\lambda
    }where the second step is because $t\ell < 1$, and the third step follows by $t\ell \leq \frac{1-\lambda}{4\lambda}$.

    Overall, we have
    \beq{
        \nonumber
        \besp{
            \left(\alpha_1 + \frac{\alpha_2\alpha_3}{1-\alpha_4}\right)^k\leq &\left(1+t^2(\gamma^2+b^2)+\frac{4\lambda t^2(\gamma^2+b^2)}{\frac{1}{2} - \frac{1}{2}\lambda}\right)^k\\
            &\leq \exp{\left( kt^2(\gamma^2+b^2)\left( 1+\frac{8}{1-\lambda}\right) \right)}.
        }
    }This completes our proof of Lemma~\ref{lemma:stoc18_lemma_4.3}.
\end{proof}

\subsection{Proof of Theorem~\ref{thm:complexchernoff}}
\label{sec:proof_complex_chernoff}
\complexchernoff*
\begin{proof}(of Theorem~\ref{thm:complexchernoff})
    Our strategy is to adopt complexification technique~\cite{dongarra1984eigenvalue}.
    For any $d\times d$ complex Hermitian matrix $\mX$, we may write $\mX = \mY + \iu \mZ$, where $\mY$ and $\iu \mZ$ are the real and imaginary parts of $\mX$, respectively. Moreover, the Hermitian property of $\mX$~(i.e., $\mX^\ast = \mX$) implies that (1) $\mY$ is real and symmetric~(i.e., $\mY^\top = \mY$); (2) $\mZ$ is real and skew symmetric~(i.e., $\mZ=-\mZ^\top$). The eigenvalues of $\mX$ can be found via a $2d \times 2d$ real symmetric matrix
    $\scriptsize \mH\triangleq \begin{bmatrix} \mY & \mZ \\ -\mZ & \mY \end{bmatrix}\normalsize$,
    where the symmetry of $\mH$ follows by the symmetry of $\mY$ and skew-symmetry of $\mZ$. Note the fact that, if the eigenvalues~(real) of $\mX$ are $\lambda_1, \lambda_2, \cdots \lambda_d$, then those of $\mH$ are $\lambda_1, \lambda_1, \lambda_2,\lambda_2, \cdots, \lambda_d, \lambda_d$.
    I.e., $\mX$ and $\mH$ have the same eigenvalues, but with different multiplicity.
    
    Using the above technique, we can formally prove Theorem~\ref{thm:complexchernoff}.
    For any complex matrix function $f: [N]\rightarrow \mathbb{C}^{d\times d}$ in Theorem~\ref{thm:complexchernoff}, we can separate its real and imaginary parts by $f = f_1 + \iu f_2$. Then we construct a real-valued matrix function
    $g: [N]\rightarrow \mathbb{R}^{2d\times 2d}$ s.t. $\forall v\in[N]$,
    $\scriptsize g(v) = \begin{bmatrix}
    f_1(v) &  f_2(v) \\
    -f_2(v) & f_1(v)
    \end{bmatrix}\normalsize$.
    According to the complexification technique, we know that (1) $\forall v\in [N], g(v)$ is real symmetric and $\norm{g(v)}{2} = \norm{f(v)}{2} \leq 1$; (2) $\sum_{v\in [N]} \pi_v g(v) = 0$. Then
    \beq{\nonumber
    \mathbb{P}\left[\lambda_{\max}\left( \frac{1}{k}\sum_{j=1}^k f(v_j)\right)\geq \epsilon\right] = \mathbb{P}\left[\lambda_{\max}\left( \frac{1}{k}\sum_{j=1}^k g(v_j)\right)\geq \epsilon\right] \leq 4\norm{\vphi}{\vpi}d^{2}\exp{\left( -(\eps^2 (1-\lambda)k / 72) \right)},
    }where the first step follows by the fact that $\frac{1}{k}\sum_{j=1}^k f(v_j)$ and $\frac{1}{k}\sum_{j=1}^k g(v_j)$ have the same eigenvalues~(with different multiplicity), 
    and the second step
    follows by Theorem~\ref{thm:chernoff}.\footnote{The additional factor 4 is because the constructed $g(v)$ has shape $2d\times 2d$.}
    The bound on $\lambda_{\min}$ also follows similarly.
    \end{proof}

\end{document}